\documentclass[journal,twoside,web]{ieeecolor}
\usepackage{generic}
\usepackage{cite}
\usepackage{amsmath,amssymb,amsfonts}
\usepackage{array}
\usepackage[caption=false,font=normalsize,labelfont=sf,textfont=sf]{subfig}
\usepackage{textcomp}
\usepackage{stfloats}
\usepackage{url}
\usepackage{verbatim}
\usepackage{graphicx}
\usepackage{placeins}
\usepackage{multirow}
\usepackage{threeparttable}
\usepackage{hyperref}
\hypersetup{hidelinks=true}

\def\BibTeX{{\rm B\kern-.05em{\sc i\kern-.025em b}\kern-.08em
    T\kern-.1667em\lower.7ex\hbox{E}\kern-.125emX}}
\begin{document}

\title{DSAINet: An Efficient Dual-Scale Attentive Interaction Network for General EEG Decoding}

\author{
    Zhiyuan Ma, \IEEEmembership{Graduate Student Member,~IEEE},
    Zeyuan Li,
    Zihao Qiu,
    Jinhao Li,
    Lingqin Meng, \\
    Xinche Zhang,
    Yixuan Liu,
    Xinke Shen, \IEEEmembership{Member,~IEEE},
    Sen Song
    \thanks{This work was supported by the National Natural Science Foundation of China under Grant 2025ZD0215701, Grant T2341003, Grant U2336214, in part by the Beijing Natural Science Foundation under Grant L257019, in part by the General Program of Guangdong Natural Science Foundation under Grant 2026A1515010121, and in part by the Shenzhen Science and Technology Innovation Committee under Grant RCBS20231211090748082. (Zhiyuan Ma, Zeyuan Li and Zihao Qiu contributed equally to this work.) (Corresponding authors: Sen Song; Xinke Shen.)}
    \thanks{Zhiyuan Ma, Zeyuan Li, Zihao Qiu, Xinche Zhang, Yixuan Liu and Sen Song are with the School of Biomedical Engineering, Tsinghua Laboratory of Brain and Intelligence, and IDG/McGovern Institute for Brain Research, Tsinghua University, Beijing 100084, China (e-mail: ma-zy25@mails.tsinghua.edu.cn; zeyuan-l23@mails.tsinghua.edu.cn; qiuzh23@mails.tsinghua.edu.cn; zhang-xc24@mails.tsinghua.edu.cn; liuyixua22@mails.tsinghua.edu.cn; songsen@tsinghua.edu.cn).}
    \thanks{Jinhao Li is with the School of Basic Medical Sciences and Tsinghua Laboratory of Brain and Intelligence, Tsinghua University, Beijing 100084, China (e-mail: jh-li23@mails.tsinghua.edu.cn).}
    \thanks{Lingqin Meng is with the Department of Psychological and Cognitive Sciences, Tsinghua University, Beijing 100084, China (e-mail: menglq24@mails.tsinghua.edu.cn).}
    \thanks{Xinke Shen is with the Department of Biomedical Engineering, Southern University of Science and Technology, Shenzhen 518055, China (e-mail: shenxk@sustech.edu.cn).}
}

\maketitle

\begin{abstract}
In real-world applications of noninvasive electroencephalography (EEG), specialized decoders often show limited generalizability across diverse tasks under subject-independent settings. One central challenge is that task-relevant EEG signals often follow different temporal organization patterns across tasks, while many existing methods rely on task-tailored architectural designs that introduce task-specific temporal inductive biases. This mismatch makes it difficult to adapt temporal modeling across tasks without changing the model configuration. To address these challenges, we propose DSAINet, an efficient dual-scale attentive interaction network for general EEG decoding. Specifically, DSAINet constructs shared spatiotemporal token representations from raw EEG signals and models diverse temporal dynamics through parallel convolutional branches at fine and coarse scales. The resulting representations are then adaptively refined by intra-branch attention to emphasize salient scale-specific patterns and by inter-branch attention to integrate task-relevant features across scales, followed by adaptive token aggregation to yield a compact representation for prediction. Extensive experiments on five downstream EEG decoding tasks across ten public datasets show that DSAINet consistently outperforms 13 representative baselines under strict subject-independent evaluation. Notably, this performance is achieved using the same architecture hyperparameters across datasets. Moreover, DSAINet achieves a favorable accuracy-efficiency trade-off with only about 77K trainable parameters and provides interpretable neurophysiological insights. The code is publicly available at https://github.com/zy0929/DSAINet.
\end{abstract}


\section{Introduction}
\IEEEPARstart{N}{oninvasive} electroencephalography (EEG) is widely used in applications such as brain–computer interfaces \cite{nicolas2012brain}, clinical diagnostics \cite{biasiucci2019electroencephalography}, and cognitive monitoring \cite{berka2007eeg}. In this work, we define general EEG decoding as the use of a unified model across multiple EEG tasks under subject-independent evaluation, with a shared architecture and consistent architecture-related hyperparameters. This objective is practically important because it tests whether a model can preserve effective feature extraction, stable performance, and cross-task transferability without relying on dataset-specific redesign \cite{kostas2021bendr, EEGNet, LMDA-Net}.

Achieving general EEG decoding in practice remains challenging. EEG tasks differ markedly in experimental paradigms and signal characteristics \cite{abiri2019comprehensive, li2025estformer, EEGNet}, and their discriminative information may be distributed across different temporal scales \cite{nicolas2012brain}. The challenge becomes even greater under subject-independent evaluation, where pronounced inter-subject variability can further alter the temporal and spatial manifestation of task-relevant EEG activity and make stable feature learning more difficult \cite{chen2025quantifying, saha2020intra, huang2023discrepancy}. Therefore, a unified EEG decoder is expected to remain robust to heterogeneous temporal structures and unseen individuals without extensive dataset-specific architectural redesign.

\begin{figure}[t]
    \centering
    \includegraphics[width=1\linewidth, trim={30 25 30 25},clip]{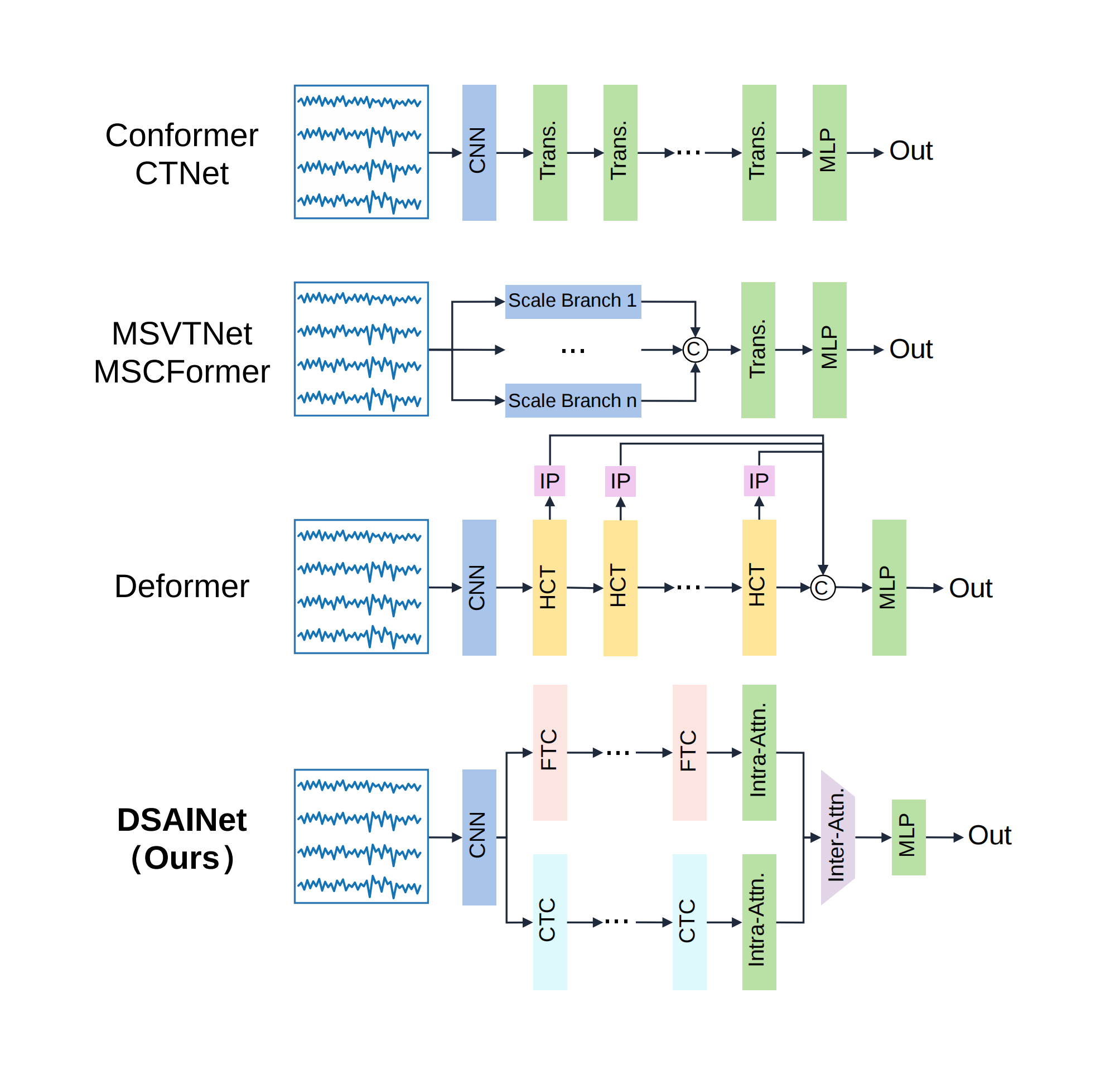}
    \captionsetup{skip=2pt}
    \caption{Architectural comparison of representative CNN-Transformer hybrid models for EEG decoding, together with the proposed DSAINet.}
    \label{fig:comparison}
\end{figure}

Existing EEG decoding methods have substantially advanced temporal modeling through more expressive convolutional, attention-based, and hybrid architectures \cite{wang2026integrating, hossain2023status}. In convolutional models, temporal patterns can be captured efficiently through local filtering and enlarged receptive fields, and multi-branch or multi-scale designs have further improved the ability to model EEG dynamics over different temporal ranges. \cite{ADFCNN, li2023parallel}. Attention-based and CNN-Transformer hybrid models have extended this line of work by enabling more flexible feature interaction, long-range dependency modeling, and adaptive representation refinement. \cite{Conformer, TMSA-Net, MSVTNet, MSCFormer}. These developments have clearly improved the modeling capacity of EEG decoders and broadened their applicability across tasks. However, from the perspective of general EEG decoding, a more fundamental limitation remains in how these architectures organize temporal modeling. In many existing designs, multi-scale temporal feature extraction and the adaptive coordination of scale-specific information are not explicitly distinguished. As a result, cross-scale information is often combined through relatively simple fusion strategies after independent extraction, or attention is required to simultaneously handle multiple modeling roles, including temporal dependency modeling, feature refinement, and feature fusion, which may increase architectural complexity and computational cost \cite{DBConformer, xiong2026ct, Deformer}. Consequently, achieving both efficient temporal modeling and adaptive cross-scale coordination within a compact unified architecture remains challenging for current EEG decoders.

To address these limitations, we propose DSAINet, an efficient dual-scale attentive interaction network for general EEG decoding. The core design principle of DSAINet is to explicitly decouple two modeling roles within a compact unified architecture: efficient extraction of temporal patterns at different scales and adaptive coordination of scale-specific information. Following this principle, DSAINet first converts raw EEG into shared spatiotemporal tokens, and then uses two temporal branches with different receptive fields to capture fine- and coarse-scale patterns within a common backbone. Each branch is implemented with stacked lightweight temporal convolution blocks based on depthwise temporal convolution and grouped pointwise transformation, so longer-range temporal dependencies are modeled mainly through compact convolutions rather than deep attention stacks. On top of these branch-wise representations, attention modules are introduced in a structured manner to serve complementary functions. Intra-branch attention performs scale-specific refinement by selectively strengthening informative responses within each branch, whereas inter-branch attention explicitly promotes information exchange across the two scales. Finally, adaptive token aggregation produces a compact representation for prediction, avoiding the need to flatten all tokens into a heavy classifier head. In this way, DSAINet provides a more balanced design for multi-scale temporal modeling and feature interaction under consistent architecture-related parameters across tasks. Fig.~\ref{fig:comparison} further illustrates this design by comparing DSAINet with representative CNN-Transformer hybrid models from an architectural perspective.

The main contributions can be summarized as follows:
\begin{itemize}
    \item We propose DSAINet, an efficient dual-scale attentive interaction network for general EEG decoding, which captures temporal patterns at fine and coarse scales.
    
    \item We develop a structured attention mechanism that separates scale-specific refinement from cross-scale integration, enabling adaptive refinement and integration of task-relevant temporal features across heterogeneous tasks.

    \item Comprehensive experiments on five EEG paradigms across ten public datasets under subject-independent evaluation demonstrate that DSAINet consistently outperforms 13 representative baselines, while also achieving a favorable accuracy–efficiency trade-off and providing meaningful interpretability.
\end{itemize}

\section{Related Work}

\subsection{CNN-based EEG Decoding}
CNN-based EEG decoding established an important foundation for end-to-end learning from raw EEG signals. Early studies such as DeepConvNet and ShallowConvNet \cite{SCNNDCNN} showed that discriminative EEG patterns can be learned directly from raw inputs, thereby reducing reliance on handcrafted feature engineering. EEGNet \cite{EEGNet} further improved compactness by introducing depthwise and separable convolutions for efficient temporal-spatial pattern learning. Later CNN-based studies further improved temporal modeling by enabling the extraction of features over different temporal ranges and the capture of richer temporal dynamics. For example, Sakhavi et al. \cite{Sakhavi-CNN} combined an FBCSP-inspired temporal representation with CNN-based learning for motor imagery decoding, whereas CP-MixedNet \cite{CP-MixedNet} used mixed-scale convolutions to capture multi-scale temporal features with controlled parameter growth. Other studies extended CNNs to more challenging settings. Kwon et al. \cite{Kwon-CNN} developed a subject-independent CNN framework to reduce calibration dependence, whereas Wang et al. \cite{Wang-CNN} proposed a weighted multi-branch structure to better exploit multisubject data for subject-specific classification. More recent lightweight CNN designs, such as MFRC-Net \cite{MFRC-Net}, further incorporated residual multi-scale temporal modeling while emphasizing parameter efficiency. Despite these advances, most CNN-based EEG decoders still focus primarily on feature extraction at different temporal ranges, whereas explicit modeling of task-relevant scale selection and cross-scale interaction remains limited.

\subsection{Attention-based EEG Decoding}
Early attention-based EEG studies primarily introduced attention as auxiliary modules for selective feature enhancement. Representative studies included Tao et al. \cite{Tao-attn}, who combined channel-wise attention with self-attention for EEG-based emotion recognition, MSCTANN \cite{MSCTANN}, which coupled multi-scale feature extraction with channel-temporal attention for motor imagery decoding, and Ying et al. \cite{Ying-attn}, who proposed a functional connectivity-based model with a lightweight attention mechanism for depression recognition. More recent studies have begun to use attention in a more structured manner beyond simple feature reweighting. For example, DAEST \cite{DAEST} estimated dynamic attention weights over spatiotemporal EEG components to model state transitions for cross-subject emotion recognition. Attention has also been incorporated more deeply into model backbones, especially in CNN-Transformer hybrids. CTNet \cite{CTNet} combined convolutional encoders with Transformer blocks to jointly capture local and global patterns for motor imagery classification, Ahn et al. \cite{Ahn-attn} proposed a multiscale convolutional Transformer that applied attention across spatial, spectral, and temporal domains for mental
imagery classification, and DBConformer \cite{DBConformer} adopted a dual-branch convolutional Transformer to model temporal dependencies and spatial interactions in parallel for EEG decoding. However, existing attention-based EEG decoders still face a practical dilemma: lightweight attention modules often offer limited task adaptability, whereas Transformer-style backbones usually assign multiple modeling roles to attention, which increase architectural complexity and computational cost.

\subsection{General EEG Decoding}
Under our definition of general EEG decoding, existing studies have mainly demonstrated broader architectural applicability across paradigms or tasks, rather than systematic validation within a unified subject-independent evaluation framework. EEGNet \cite{EEGNet} was an early representative attempt, showing that a compact network could be applied to different BCI paradigms within a relatively unified architecture. Conformer \cite{Conformer} extended this line of research through a CNN-Transformer hybrid that combines convolutional embedding for local pattern extraction with self-attention for global dependency modeling. LMDA-Net \cite{LMDA-Net} was proposed as a lightweight multi-dimensional attention network for general EEG-based BCIs, using channel and depth attention to integrate EEG representations across different dimensions. More recent studies have adopted increasingly structured architectures to improve broader applicability. For example, Deformer \cite{Deformer} introduced coarse-to-fine temporal modeling to capture hierarchical EEG dynamics, whereas DBConformer \cite{DBConformer} employed a dual-branch convolutional Transformer to separately model temporal dependencies and inter-channel interactions. Taken together, these studies mark important progress toward general EEG decoding, but robust and efficient modeling of heterogeneous task-related EEG dynamics within a shared architectural configuration remains an open problem.

\section{Method}
\label{sec:method}

\begin{figure*}[!t]
    \centering
    \includegraphics[width=\linewidth]{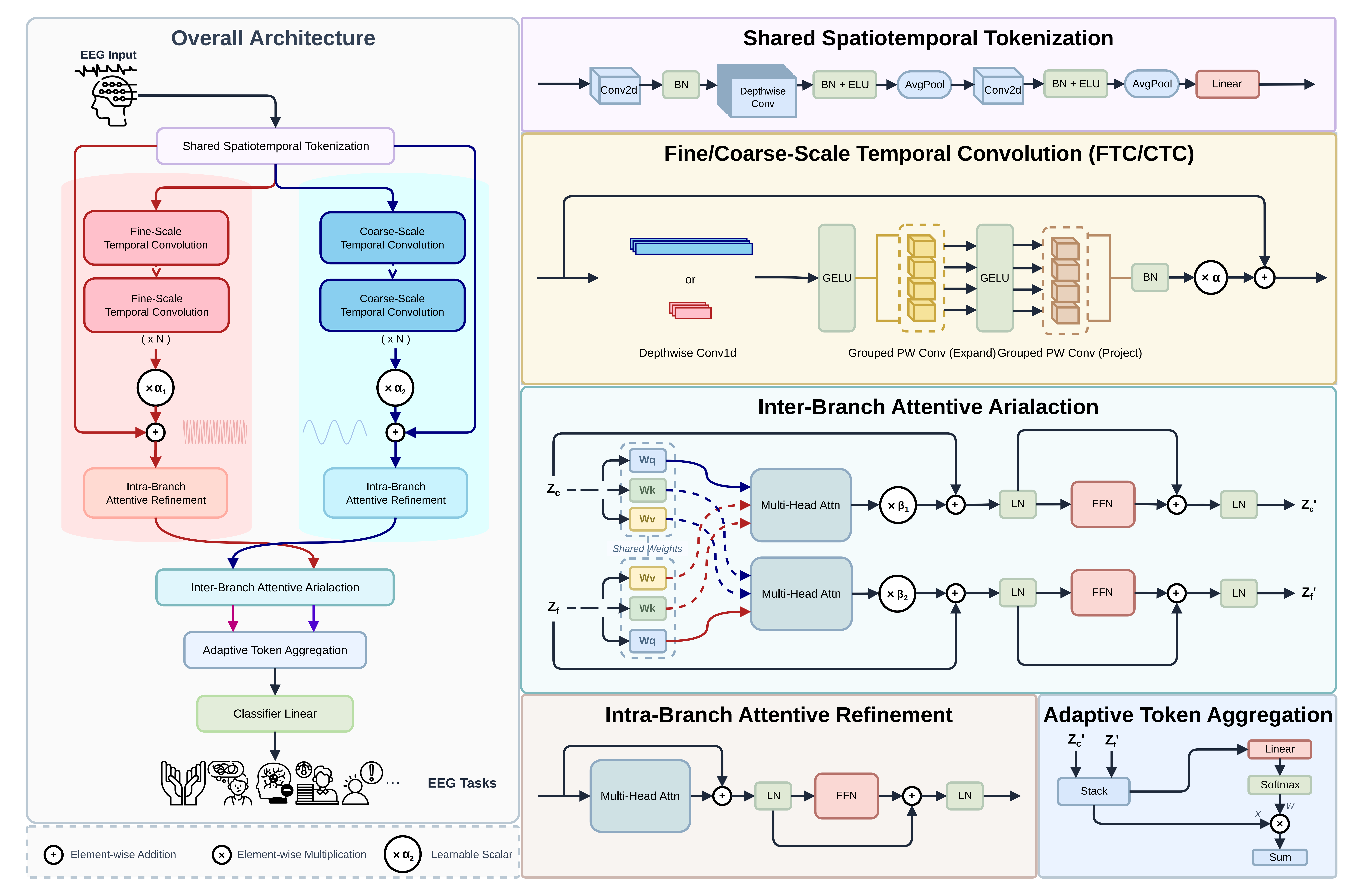}
    \caption{Overall architecture of DSAINet. The model first converts raw EEG into shared spatiotemporal tokens, then captures fine- and coarse-scale temporal patterns through two parallel temporal convolution branches, with the resulting branch-wise features further refined by intra-branch attention, integrated by inter-branch attention, and finally aggregated by adaptive token pooling for classification.}
    \label{fig:architecture}
\end{figure*}

The proposed DSAINet follows a progressive feature modeling pipeline. Shared Spatiotemporal Tokenization first converts raw EEG into compact spatiotemporal tokens. These tokens are then processed by Fine/Coarse-Scale Temporal Convolution (FTC/CTC), which uses two parallel branches with different temporal receptive fields to capture fine- and coarse-scale temporal features. On top of the resulting branch features, Intra-Branch Attentive Refinement performs branch-wise feature refinement, and Inter-Branch Attentive Interaction further enables explicit information exchange across scales. The resulting representations are finally aggregated by Adaptive Token Aggregation for prediction.

\subsection{Shared Spatiotemporal Tokenization}
Let an EEG trial be denoted by
\begin{equation}
\mathbf{X} \in \mathbb{R}^{C \times T},
\end{equation}
where $C$ and $T$ denote the number of channels and temporal samples, respectively. We first employ a lightweight CNN tokenizer to transform the raw EEG into a sequence of shared spatiotemporal tokens. Specifically, the tokenizer contains two temporal convolutions, a depthwise spatial convolution spanning all channels, and two temporal downsampling stages:
\begin{equation}
\mathbf{F} = \Phi_{\mathrm{tok}}(\mathbf{X}) \in \mathbb{R}^{f_2 \times 1 \times N},
\end{equation}
where $\Phi_{\mathrm{tok}}(\cdot)$ denotes the tokenization module, $f_2$ is the output channel dimension, and $N$ is the number of temporal tokens after pooling.

We then squeeze the singleton spatial dimension and rearrange the feature map into token form:
\begin{equation}
\tilde{\mathbf{Z}} = \mathrm{Transpose}(\mathrm{Squeeze}(\mathbf{F})) \in \mathbb{R}^{N \times f_2}.
\end{equation}
A linear projection is used to map the tokens into a unified embedding space of dimension $d$:
\begin{equation}
\mathbf{Z}^{(0)} = \mathcal{P}(\tilde{\mathbf{Z}})\in\mathbb{R}^{N\times d},
\end{equation}
where $\mathcal{P}(\cdot)$ is an identity mapping when $f_2=d$, and is otherwise implemented as a learnable linear layer. Following the implementation, the projected tokens are scaled and combined with learnable positional embeddings:
\begin{equation}
\mathbf{Z}_0 = \mathrm{PE}\!\left(\sqrt{d}\,\mathbf{Z}^{(0)}\right)\in\mathbb{R}^{N\times d}.
\end{equation}
Here, $\mathrm{PE}(\cdot)$ denotes learnable positional encoding, which preserves token order for the subsequent attentive modules.

\subsection{Fine/Coarse-Scale Temporal Convolution}
To model EEG dynamics at different temporal resolutions, DSAINet uses two parallel temporal branches that share the same token input $\mathbf{Z}_0$ but adopt different kernel sizes. The fine-scale branch is designed to emphasize local temporal details, while the coarse-scale branch captures broader contextual dependencies. Before temporal convolution, the token sequence is transposed into channel-first format:
\begin{equation}
\mathbf{H}_0 = \mathbf{Z}_0^\top \in \mathbb{R}^{d\times N}.
\end{equation}

Each branch is composed of stacked lightweight temporal convolution modules. For branch $b \in \{f,c\}$ and layer $\ell$, the module can be written as
\begin{equation}
\tilde{\mathbf{H}}_{b}^{(\ell)}
=
\mathrm{BN}\!\left(
\mathrm{PW}_{b,2}^{(\ell)}
\Big(
\sigma\big(
\mathrm{PW}_{b,1}^{(\ell)}
(
\sigma(
\mathrm{DW}_{k_b^{(\ell)}}(\mathbf{H}_{b}^{(\ell-1)})
)
)\big)
\Big)
\right),
\end{equation}
where $\mathrm{DW}_{k_b^{(\ell)}}(\cdot)$ denotes depthwise temporal convolution with kernel size $k_b^{(\ell)}$, $\mathrm{PW}_{b,1}^{(\ell)}(\cdot)$ and $\mathrm{PW}_{b,2}^{(\ell)}(\cdot)$ denote grouped pointwise transformations for channel expansion and projection, $\sigma(\cdot)$ is the GELU activation, and $\mathrm{BN}(\cdot)$ is batch normalization. The branch output is updated through a residual connection with a learnable scaling factor:
\begin{equation}
\mathbf{H}_{b}^{(\ell)}
=
\mathbf{H}_{b}^{(\ell-1)}+\alpha_{b}^{(\ell)}\tilde{\mathbf{H}}_{b}^{(\ell)}.
\end{equation}
After the stacked temporal modules, the two branches are rearranged back to token form:
\begin{equation}
\mathbf{Z}_{f}=\left(\mathbf{H}_{f}^{(L_f)}\right)^\top,\qquad
\mathbf{Z}_{c}=\left(\mathbf{H}_{c}^{(L_c)}\right)^\top,
\end{equation}
where $\mathbf{Z}_{f},\mathbf{Z}_{c}\in\mathbb{R}^{N\times d}$.

In the model design shown in Fig.~\ref{fig:architecture}, the shared tokens can also be injected back into the two branches through learnable residual scalars:
\begin{equation}
\mathbf{Z}_{f}\leftarrow \mathbf{Z}_{f}+\alpha_{1}\mathbf{Z}_{0},\qquad
\mathbf{Z}_{c}\leftarrow \mathbf{Z}_{c}+\alpha_{2}\mathbf{Z}_{0}.
\end{equation}
This operation helps preserve the shared token information while allowing the two branches to specialize under different receptive fields.

\subsection{Intra-Branch Attentive Refinement}
After temporal convolution, each branch is refined independently by a Transformer-style self-attention block. This stage aims to strengthen salient scale-specific token relationships within each branch. For branch $b\in\{f,c\}$ at attentive layer $m$, the self-attention refinement is formulated as
\begin{equation}
\bar{\mathbf{Z}}_{b}^{(m)}
=
\mathrm{LN}\!\left(
\mathbf{Z}_{b}^{(m-1)}+
\mathrm{MHA}\!\left(
\mathbf{Z}_{b}^{(m-1)},
\mathbf{Z}_{b}^{(m-1)},
\mathbf{Z}_{b}^{(m-1)}
\right)
\right),
\end{equation}
\begin{equation}
\mathbf{Z}_{b}^{(m)}
=
\mathrm{LN}\!\left(
\bar{\mathbf{Z}}_{b}^{(m)}+\mathrm{FFN}\!\left(\bar{\mathbf{Z}}_{b}^{(m)}\right)
\right),
\end{equation}
where $\mathrm{MHA}(\cdot)$ denotes multi-head self-attention, $\mathrm{FFN}(\cdot)$ denotes a two-layer feed-forward network with GELU nonlinearity, and $\mathrm{LN}(\cdot)$ denotes layer normalization. Unlike generic self-attention usage, here the role of intra-branch attention is not to replace temporal modeling itself, but to refine the branch-wise features that have already been extracted at a specific temporal scale.

\subsection{Inter-Branch Attentive Interaction}
DSAINet introduces a symmetric cross-attention module to exchange information between fine- and coarse-scale representations. Given the refined branch features, the cross-scale interaction is defined as
\begin{equation}
\bar{\mathbf{Z}}_{f}
=
\mathrm{LN}\!\left(
\mathbf{Z}_{f}
+
\beta_{1}\,
\mathrm{MHA}\!\left(
\mathbf{Z}_{f},
\mathbf{Z}_{c},
\mathbf{Z}_{c}
\right)
\right),
\end{equation}
\begin{equation}
\mathbf{Z}_{f}^{\prime}
=
\mathrm{LN}\!\left(
\bar{\mathbf{Z}}_{f}
+
\mathrm{FFN}\!\left(\bar{\mathbf{Z}}_{f}\right)
\right),
\end{equation}
\begin{equation}
\bar{\mathbf{Z}}_{c}
=
\mathrm{LN}\!\left(
\mathbf{Z}_{c}
+
\beta_{2}\,
\mathrm{MHA}\!\left(
\mathbf{Z}_{c},
\mathbf{Z}_{f},
\mathbf{Z}_{f}
\right)
\right),
\end{equation}
\begin{equation}
\mathbf{Z}_{c}^{\prime}
=
\mathrm{LN}\!\left(
\bar{\mathbf{Z}}_{c}
+
\mathrm{FFN}\!\left(\bar{\mathbf{Z}}_{c}\right)
\right),
\end{equation}
where $\beta_{1}$ and $\beta_{2}$ are learnable interaction strengths. In this way, the fine-scale branch can selectively absorb broader contextual cues from the coarse-scale branch, while the coarse-scale branch can also incorporate more local discriminative details from the fine-scale branch. This explicit bidirectional interaction is a core part of the proposed dual-scale collaborative design.

\subsection{Adaptive Token Aggregation and Classification}
After attentive refinement and interaction, the two branch outputs $\mathbf{Z}_{f}^{\prime}$ and $\mathbf{Z}_{c}^{\prime}$ are aggregated into compact branch-level representations. Instead of average pooling, DSAINet adopts adaptive token aggregation to assign larger weights to more informative tokens. For branch $b\in\{f,c\}$ and token $n$, the token weight is computed as
\begin{equation}
w_{b,n}=
\frac{\exp\!\left(\mathbf{q}^{\top}\mathbf{z}_{b,n}\right)}
{\sum_{j=1}^{N}\exp\!\left(\mathbf{q}^{\top}\mathbf{z}_{b,j}\right)},
\end{equation}
where $\mathbf{z}_{b,n}\in\mathbb{R}^{d}$ is the $n$-th token of branch $b$, and $\mathbf{q}\in\mathbb{R}^{d}$ is a learnable projection vector shared across branches in the implementation. The pooled branch representation is then
\begin{equation}
\mathbf{p}_{b}=\sum_{n=1}^{N}w_{b,n}\mathbf{z}_{b,n}\in\mathbb{R}^{d}.
\end{equation}
Finally, the pooled fine- and coarse-scale features are concatenated and passed to a lightweight classifier:
\begin{equation}
\hat{\mathbf{y}}=\mathbf{W}_{\mathrm{cls}}[\mathbf{p}_{f}\,\|\,\mathbf{p}_{c}]+\mathbf{b}_{\mathrm{cls}},
\end{equation}
where $\hat{\mathbf{y}}$ denotes the prediction logits. This design avoids flattening all tokens into a heavy classification head and thereby preserves the compactness of the overall model.

\section{Experiment}

\subsection{Datasets}
The experiments were conducted on five EEG decoding paradigms: motor imagery (MI), mental disorder, neurodegenerative disorder, mental workload, and cognitive attention. Table~\ref{tab:dataset_summary} provides a summary of the datasets used in this study.

\begin{table*}[!htbp]
    \caption{Summary of datasets and task types across different paradigms.}
    \label{tab:dataset_summary}
    \renewcommand{\arraystretch}{1.3}
    \centering
    \footnotesize
    \setlength{\tabcolsep}{4pt}
    \begin{threeparttable}
    \begin{tabular}{ccccccccccc}
        \hline
            Paradigm & Dataset & Subjects & Channels & Sessions & \begin{tabular}[c]{@{}c@{}}Sampling\\Rate (Hz)\end{tabular} & \begin{tabular}[c]{@{}c@{}}Trial Length\\(s)\end{tabular} & Overlap & Total Trials & Classes & Task Types\\
        \hline
        \multirow{5}{*}{Motor Imagery} 
        & BCIC-IV-2a & 9 & 22 & 2 & 250 & 4 & 0\% & 5,184 & 4 & LH/RH/F/T \\
        & BCIC-IV-2b & 9 & 3 & 5 & 250 & 4 & 0\% & 6,520 & 2 & LH/RH \\
        & Zhou2016 & 4 & 14 & 3 & 200 & 5 & 0\% & 1,800 & 3 & LH/RH/F \\
        & OpenBMI & 54 & 20 & 2 & 250 & 4 & 0\% & 21,600 & 2 & LH/RH \\
        & PhysioNet-MI & 109 & 64 & 1 & 250 & 4 & 0\% & 9,837 & 4 & LFi/RFi/BFi/F \\
        \hline
        \multirow{1}{*}{Mental Disorder} 
        & Mumtaz2017 & 63 & 19 & 1-2 & 200 & 5 & 0\% & 7,143 & 2 & HC/MDD \\
        \hline
        \multirow{2}{*}{Neurodegenerative Disorder} 
        & ADFTD & 88 & 19 & 1 & 250 & 4 & 0\% & 17,419 & 3 & HC/AD/FTD \\
        & Rockhill2021 & 31 & 32 & 1 & 250 & 4 & 50\% & 2,931 & 2 & HC/PD \\
        \hline
        \multirow{1}{*}{Mental Workload} 
        & EEGMat & 36 & 19 & 1 & 500 & 2 & 50\% & 4,248 & 2 & RS/MA\\
        \hline
        \multirow{1}{*}{Cognitive Attention} 
        & Shin2018 & 26 & 28 & 3 & 250 & 4 & 0\% & 4,680 & 2 & RS/DSR\\
        \hline
    \end{tabular}
    \begin{tablenotes}[flushleft]
    \footnotesize\raggedright
    \item \textit{Abbreviations:} LH = left hand; RH = right hand; F = both feet; T = tongue; LFi = left fist; RFi = right fist; BFi = both fists. 
    HC = healthy controls; MDD = major depressive disorder; AD = Alzheimer's disease; FTD = frontotemporal dementia; PD = Parkinson's disease. 
    RS = resting state; MA = mental arithmetic; DSR = discrimination/selection response.
    \end{tablenotes}
    \end{threeparttable}
\end{table*}

\subsubsection{Motor Imagery Datasets}
We adopted five public MI datasets that differ in subject number and channel count.

\begin{itemize}
    \item 
    BCIC-IV-2a \cite{BCIC-IV-2a}: This dataset contains recordings from 9 subjects acquired with 22 EEG channels. Each subject performed four MI tasks, namely left-hand, right-hand, both-feet, and tongue imagery, across two sessions. In total, the dataset provides 5,184 trials, each lasting 4 s.
    
    \item 
    BCIC-IV-2b \cite{BCIC-IV-2b}: This dataset includes 9 subjects recorded with merely 3 EEG channels. The subjects performed two MI tasks, i.e., left-hand and right-hand imagery, over five sessions, yielding a collection of 6,520 trials of 4 s each.
    
    \item 
    Zhou2016 \cite{Zhou2016}: This dataset was collected from 4 subjects using 14 EEG channels. The subjects performed three MI tasks, namely left-hand, right-hand, and foot-movement imagery, in three sessions. The dataset contains 1,800 trials, each lasting 5 s.
    
    \item 
    OpenBMI \cite{OpenBMI}: This dataset contains recordings from 54 subjects. Following \cite{OpenBMI}, only 20 electrodes over the motor cortex were retained. The subjects performed left-hand and right-hand imagery tasks in two sessions, yielding 21,600 trials of 4 s each.
    
    \item 
    PhysioNet-MI \cite{PhysioNet-MI}: This dataset includes 109 subjects recorded with 64 EEG channels. A total of 9,837 MI trials corresponding to four imagery classes, namely left fist, right fist, both fists, and both feet, were used. Trials corresponding to actual motor execution were excluded.
\end{itemize}

\subsubsection{Mental Disorder Dataset}
\begin{itemize}
    \item 
    Mumtaz2017 \cite{Mumtaz2017}: This dataset consists of 34 subjects with major depressive disorder (MDD) and 30 healthy controls (HC). We retained the eyes-open (EO) and eyes-closed (EC) sessions and excluded the TASK (SSVEP) session. Not every subject had both EO and EC sessions, and one subject in the MDD group had only TASK recordings and was therefore removed. After preprocessing and segmentation, the dataset contains 7,143 trials.
\end{itemize}

\subsubsection{Neurodegenerative Disorder Datasets}
\begin{itemize}
    \item 
    ADFTD \cite{ADFTD}: This dataset includes 36 subjects with Alzheimer's disease (AD), 23 subjects with frontotemporal dementia (FTD), and 29 healthy controls (HC). After segmentation, the dataset yields 17,419 trials of 4 s each.
    
    \item 
    Rockhill2021 \cite{Rockhill2021}: This dataset contains recordings from 15 patients with Parkinson's disease (PD) and 16 healthy controls (HC), acquired with 32 EEG channels. For PD subjects, only the medication-ON session was used (where the recordings were taken after the patients had taken medicine). Owing to the limited sample size in PD, a 2-s overlap was applied during 4-s segmentation to alleviate data scarcity. In total, 2,931 trials were obtained.
\end{itemize}

\subsubsection{Mental Workload Dataset}
\begin{itemize}
    \item 
    EEGMat \cite{EEGMat, PhysioNet}: This dataset was collected from 36 subjects with 19 EEG channels before and during mental arithmetic tasks. Each mental-workload episode lasts 60 s. To balance the two classes, only the last 60 s of the resting-state EEG were used as the low-workload condition, following \cite{Deformer}. To alleviate data scarcity, the signals were segmented into 2-s epochs with 1-s overlap, yielding a total of 4,248 trials.
\end{itemize}

\subsubsection{Cognitive Attention Dataset}
\begin{itemize}
    \item 
    Shin2018 \cite{Shin2018}: This dataset includes 26 subjects recorded with 28 EEG channels. The subjects performed the discrimination/selection response (DSR) task in three sessions. Following \cite{Deformer}, only the first half of each attention episode (20 s) was retained to balance the classes, yielding 4,680 trials of 4 s each.
\end{itemize}

\subsubsection{Data Preprocessing}
We applied a standard EEG preprocessing pipeline, including filtering, resampling and epoch segmentation.  For the MI datasets, as well as Rockhill2021 and Shin2018, a band-pass filter of 0.5-40 Hz was used. For ADFTD and EEGMat, the band-pass range was set to 0.5-45 Hz. For Mumtaz2017, we used a 0.3-75 Hz band-pass filter and additionally applied a 50 Hz notch filter to suppress power-line interference. The resampling settings and trial lengths are summarized in Table~\ref{tab:dataset_summary}. After preprocessing, we applied channel-wise z-score normalization. 

\subsection{Baselines}
We compared our proposed DSAINet with thirteen EEG decoding networks in the experiments:

\begin{itemize}
    \item 
    ShallowConvNet \cite{SCNNDCNN}: a compact CNN inspired by filter bank common spatial pattern, using temporal and spatial convolutions to extract discriminative band-power features from raw signals.
    \item 
    DeepConvNet \cite{SCNNDCNN}: a deeper CNN that stacks multiple convolutional blocks to learn more complex EEG representations directly from raw signals.
    \item 
    EEGNet \cite{EEGNet}: a compact CNN with depthwise separable convolution to reduce learnable parameters while retaining robust spatiotemporal representations.
    \item 
    ADFCNN \cite{ADFCNN}: a dual-branch CNN with temporal-spatial convolutions of different kernel sizes to simulate different receptive fields, where features from the two branches are fused with an attention mechanism before classification.
    \item 
    Conformer \cite{Conformer}: a compact convolutional Transformer that combines CNNs with a self-attention module to jointly learn local features and global dependencies from raw EEG signals.
    \item 
    LMDA-Net \cite{LMDA-Net}: a lightweight multi-dimensional attention network that combines a shallow convolutional backbone with channel and depth attention modules to better integrate temporal, spatial, and high-dimensional features.
    \item 
    MSVTNet \cite{MSVTNet}: a CNN-Transformer model that uses parallel multi-scale convolutional branches to extract local features, then applies a transformer to capture cross-scale interactions and global temporal dependencies.
    \item 
    CTNet \cite{CTNet}: a convolutional Transformer that first employs temporal and spatial convolutions to extract local spatiotemporal patterns, and then uses a Transformer encoder to model global dependencies for classification.
    \item 
    TMSA-Net \cite{TMSA-Net}: an efficient CNN-Transformer that incorporates multi-scale convolution and a temporal multi-scale attention mechanism within self-attention to jointly capture local and global dependencies.
    \item 
    MGFormer \cite{MGFormer}: a CNN-Transformer model using multi-granular token encoding and hybrid feature fusion to jointly capture temporal, spatial, and spectral features.
    \item 
    Deformer \cite{Deformer}: a dense convolutional Transformer that combines a shallow CNN encoder, hierarchical coarse-to-fine temporal modeling, and dense information purification to capture multi-level EEG dynamics.
    \item
    DBConformer \cite{DBConformer}: a dual-branch convolutional Transformer that models temporal dynamics and spatial inter-channel relationships in parallel, and further refines spatial representations with a lightweight channel attention module before fusion and classification.
    \item 
    MSCFormer \cite{MSCFormer}: a CNN-Transformer hybrid that uses parallel multi-scale convolution branches to extract local spatiotemporal features and an attention-based encoder to capture global dependencies for classification.
\end{itemize}

\subsection{Experiment Settings}
In this study, we focused exclusively on subject-independent evaluation, where the training, validation, and test sets contained disjoint subjects, providing a more realistic assessment of generalization to unseen individuals. To this end, we used two subject-independent protocols: leave-one-subject-out (LOSO) and subject-level k-fold cross-validation.

For BCIC-IV-2a, BCIC-IV-2b, and Zhou2016, we employed the LOSO protocol due to their relatively limited numbers of subjects. In each LOSO run, one subject was held out for testing, and all remaining subjects were used for model development. Following \cite{Deformer}, for each training subject, 20\% of samples were further reserved for validation via label-stratified splitting, with the remaining 80\% used for training.

For the remaining datasets, we employed subject-level k-fold cross-validation, in which subjects were partitioned into k folds, with stratification applied where appropriate to preserve label balance. In each run, one fold served as the test set, one of the remaining folds was randomly selected as the validation set, and the rest were used for training. We used 10-fold cross-validation for all datasets except Rockhill2021, for which 5-fold cross-validation was adopted due to the limited number of subjects and substantial inter-subject variability.

\subsection{Implementation Details}
All experiments were conducted on a server with eight NVIDIA GeForce RTX 4090 GPUs. Across all datasets, the architectural hyperparameters were fixed, while the dataset-specific training hyperparameters were summarized in Table~\ref{tab:training}.

Specifically, the embedding dimension was 40 and the number of attention heads was 4. In the Shared Spatiotemporal Tokenization stage, $f_1$ was 16, the temporal kernel size were 64 and 16, the depth multiplier $D$ was 2, and the pooling sizes were 4 and 8. This stage was followed by two temporal convolution branches for coarse- and fine-scale modeling. In the coarse-scale branch, the two stacked temporal convolution blocks used kernel sizes of 11 and 15, respectively, whereas those in the fine-scale branch used kernel sizes of 3 and 7. In both branches, the convolution expansion ratio was set to 4. The resulting branch features were then refined by the Intra-Branch Attentive Refinement module and fused by the Inter-Branch Attentive Interaction module. In both attentive modules, the feed-forward expansion ratio was 2. Dropout was applied at a rate of 0.25 throughout the network.

For optimization, we used the Adam optimizer with a weight decay of $1\times10^{-4}$. The batch size, initial learning rate, and maximum number of epochs were selected according to the scale, subject number, and heterogeneity of each dataset. In general, datasets with more training samples permitted the use of larger batch sizes, whereas datasets with fewer subjects or greater inter-subject variability required smaller learning rates and shorter training schedules to improve optimization stability and mitigate the risk of overfitting.

\subsection{Evaluation Metrics}
We use accuracy (ACC) and weighted F1-score for performance evaluation. ACC is defined as
\begin{equation}
\mathrm{ACC} = \frac{1}{N}\sum_{i=1}^{N}\mathbf{1}(\hat{y}_i = y_i),
\end{equation}
where $N$ is the number of test samples, $y_i$ is the ground-truth label, and $\hat{y}_i$ is the predicted label.

Since several datasets exhibit class imbalance, we further report the weighted F1-score:
\begin{equation}
\mathrm{weighted\text{ }F1\text{-}score}
= \sum_{c=1}^{C}\frac{N_c}{N}\cdot
\frac{2 \cdot \mathrm{Precision}_c \cdot \mathrm{Recall}_c}
{\mathrm{Precision}_c + \mathrm{Recall}_c},
\end{equation}
where $C$ is the number of classes, $N_c$ is the number of samples in class $c$, and $\mathrm{Precision}_c$ and $\mathrm{Recall}_c$ are the precision and recall of class $c$, respectively. We tested the models on five different random seeds. The performance is reported as mean $\pm$ standard deviation over all folds or LOSO runs across the five random seeds.

\begin{table}[!htbp]
\caption{Training configurations and evaluation protocols for different datasets.}
\renewcommand{\arraystretch}{1.3}
\centering
\footnotesize
\setlength{\tabcolsep}{10pt}
\begin{tabular}{lcccc}
\hline
Dataset & Batch & LR & Epochs & Protocol \\
\hline
BCIC-IV-2a & 32 & 1e-3 & 100 & LOSO \\
BCIC-IV-2b & 32 & 1e-3 & 100 & LOSO \\
Zhou2016 & 32 & 1e-3 & 100 & LOSO \\
OpenBMI & 128 & 1e-3 & 100 & 10-Fold \\
PhysioNet-MI & 128 & 1e-3 & 100 & 10-Fold \\
Mumtaz2017 & 128 & 1e-4 & 30 & 10-Fold \\
ADFTD & 128 & 1e-4 & 30 & 10-Fold \\
Rockhill2021 & 32 & 1e-4 & 30 & 5-Fold \\
EEGMat & 32 & 1e-3 & 30 & 10-Fold \\
Shin2018 & 32 & 1e-3 & 100 & 10-Fold \\
\hline
\end{tabular}
\label{tab:training}
\end{table}

\section{Results and Analyses}

\subsection{Performance Comparison}

\begin{table*}[!t]
\caption{Performance comparison on five motor imagery datasets using ACC (\%) and weighted F1-score (\%), reported as mean$\pm$std. \\Best and second-best results are highlighted in bold and underlined, respectively.}
\label{tab:mi_result}
\centering
\renewcommand{\arraystretch}{1.3}
\setlength{\tabcolsep}{2pt}
\begin{tabular}{l|cc|cc|cc|cc|cc}
\hline
\multirow{2}{*}{\centering Method}
& \multicolumn{2}{c|}{BCIC-IV-2a}
& \multicolumn{2}{c|}{BCIC-IV-2b}
& \multicolumn{2}{c|}{Zhou2016}
& \multicolumn{2}{c|}{OpenBMI}
& \multicolumn{2}{c}{PhysioNet-MI} \\
\cline{2-11}
& ACC & F1
& ACC & F1
& ACC & F1
& ACC & F1
& ACC & F1 \\
\hline

ShallowConvNet
& 55.43$\pm$1.27 & 52.96$\pm$1.36
& 75.12$\pm$0.61 & 74.58$\pm$0.71
& 74.41$\pm$1.18 & 73.91$\pm$1.17
& 79.16$\pm$0.19 & 79.06$\pm$0.20
& 58.66$\pm$0.20 & 58.69$\pm$0.28 \\

DeepConvNet
& 59.24$\pm$0.43 & 56.96$\pm$0.39
& 75.70$\pm$0.49 & 75.27$\pm$0.59
& 74.68$\pm$0.58 & 74.36$\pm$0.61
& 81.53$\pm$0.25 & 81.48$\pm$0.27
& 61.82$\pm$0.33 & 61.70$\pm$0.34 \\

EEGNet
& 58.68$\pm$0.72 & 56.41$\pm$0.98
& 75.91$\pm$0.51 & 75.33$\pm$0.61
& 75.41$\pm$1.58 & 74.94$\pm$1.64
& 80.67$\pm$0.40 & 80.63$\pm$0.40
& 60.43$\pm$0.14 & 60.48$\pm$0.12 \\

ADFCNN
& 59.95$\pm$0.81 & 57.45$\pm$0.93
& 76.22$\pm$0.32 & 75.65$\pm$0.33
& 76.65$\pm$1.33 & 76.14$\pm$1.56
& 81.32$\pm$0.51 & 81.26$\pm$0.53
& 61.54$\pm$0.15 & 61.60$\pm$0.13 \\

LMDA-Net
& 54.99$\pm$0.25 & 52.23$\pm$0.15
& 76.27$\pm$0.29 & 75.85$\pm$0.31
& 76.42$\pm$0.63 & 76.14$\pm$0.82
& 80.61$\pm$0.44 & 80.55$\pm$0.46
& 59.39$\pm$0.29 & 59.45$\pm$0.29 \\

MSVTNet
& 58.00$\pm$0.88 & 54.99$\pm$1.15
& 75.97$\pm$0.57 & 75.48$\pm$0.65
& 76.30$\pm$1.06 & 75.84$\pm$1.09
& 80.47$\pm$0.24 & 80.40$\pm$0.22
& 62.68$\pm$0.18 & 62.48$\pm$0.17 \\

CTNet
& \underline{60.08$\pm$0.71} & \underline{57.62$\pm$0.76}
& 76.30$\pm$0.56 & 75.82$\pm$0.62
& 74.41$\pm$0.55 & 73.64$\pm$0.78
& 81.89$\pm$0.36 & 81.84$\pm$0.36
& 62.67$\pm$0.17 & 62.69$\pm$0.19 \\

TMSA-Net
& 58.06$\pm$0.93 & 55.46$\pm$1.16
& 75.89$\pm$0.44 & 75.38$\pm$0.55
& 75.68$\pm$1.10 & 75.03$\pm$1.22
& 78.95$\pm$0.23 & 78.87$\pm$0.24
& 59.23$\pm$0.28 & 59.14$\pm$0.33 \\

Deformer
& 59.39$\pm$0.53 & 56.80$\pm$0.66
& 76.05$\pm$0.40 & 75.56$\pm$0.42
& 76.01$\pm$1.12 & 75.48$\pm$1.19
& \underline{81.99$\pm$0.17} & \underline{81.94$\pm$0.16}
& \underline{63.38$\pm$0.15} & \underline{63.37$\pm$0.13} \\

DBConformer
& 58.14$\pm$0.36 & 55.48$\pm$0.41
& \underline{76.37$\pm$0.46} & \underline{75.91$\pm$0.48}
& \underline{76.69$\pm$0.56} & \underline{76.29$\pm$0.58}
& 79.42$\pm$0.15 & 79.31$\pm$0.16
& 61.89$\pm$0.22 & 61.94$\pm$0.21 \\

MSCFormer
& 58.93$\pm$0.73 & 56.05$\pm$0.78
& 76.05$\pm$0.50 & 75.53$\pm$0.57
& 75.80$\pm$1.01 & 75.10$\pm$1.22
& 80.52$\pm$0.48 & 80.46$\pm$0.49
& 62.89$\pm$0.35 & 62.85$\pm$0.40 \\

\hline
\textbf{DSAINet (Ours)}
& \textbf{61.79$\pm$0.68} & \textbf{59.65$\pm$0.77}
& \textbf{76.41$\pm$0.31} & \textbf{75.91$\pm$0.36}
& \textbf{78.21$\pm$0.47} & \textbf{77.85$\pm$0.45}
& \textbf{82.89$\pm$0.13} & \textbf{82.86$\pm$0.13}
& \textbf{63.90$\pm$0.14} & \textbf{63.87$\pm$0.14} \\

\hline
\end{tabular}
\end{table*}

\begin{table*}[!t]
\caption{Performance comparison on five downstream datasets using ACC (\%) and weighted F1-score (\%), reported as mean$\pm$std. \\Best and second-best results are highlighted in bold and underlined, respectively.}
\label{tab:other_result}
\centering
\renewcommand{\arraystretch}{1.3}
\setlength{\tabcolsep}{2pt}
\begin{tabular}{l|cc|cc|cc|cc|cc}
\hline
\multirow{2}{*}{\centering Method}
& \multicolumn{2}{c|}{Mumtaz2017}
& \multicolumn{2}{c|}{ADFTD}
& \multicolumn{2}{c|}{Rockhill2021}
& \multicolumn{2}{c|}{EEGMat}
& \multicolumn{2}{c}{Shin2018} \\
\cline{2-11}
& ACC & F1
& ACC & F1
& ACC & F1
& ACC & F1
& ACC & F1 \\
\hline

ShallowConvNet
& 85.99$\pm$0.63 & 85.69$\pm$0.75
& 56.91$\pm$0.43 & 55.34$\pm$0.28
& 70.37$\pm$0.73 & 69.29$\pm$1.46
& 68.07$\pm$0.30 & 66.97$\pm$0.78
& 75.86$\pm$0.24 & 75.51$\pm$0.60 \\

DeepConvNet
& 85.70$\pm$0.60 & 85.16$\pm$0.90
& 56.00$\pm$0.91 & 53.86$\pm$1.42
& 68.59$\pm$0.74 & 66.51$\pm$1.72
& 67.50$\pm$0.26 & 66.64$\pm$0.77
& 79.95$\pm$0.45 & 79.64$\pm$0.50 \\

EEGNet
& 82.65$\pm$1.01 & 82.15$\pm$1.21
& 54.84$\pm$0.79 & 53.04$\pm$1.04
& 69.62$\pm$1.35 & 68.58$\pm$2.06
& 68.48$\pm$0.42 & 67.46$\pm$0.59
& 78.08$\pm$0.24 & 77.78$\pm$0.31 \\

ADFCNN
& 85.27$\pm$0.67 & 84.88$\pm$0.69
& 57.55$\pm$0.40 & \underline{56.60$\pm$0.51}
& 71.68$\pm$1.16 & 70.65$\pm$1.59
& 70.90$\pm$0.19 & 70.26$\pm$0.45
& 77.49$\pm$0.66 & 77.12$\pm$0.79 \\

Conformer
& \underline{87.79$\pm$0.37} & \underline{87.55$\pm$0.38}
& 56.76$\pm$0.54 & 54.98$\pm$0.91
& 70.49$\pm$1.21 & 69.68$\pm$1.11
& 69.48$\pm$0.74 & 68.53$\pm$0.87
& 79.25$\pm$0.29 & 79.03$\pm$0.29 \\

LMDA-Net
& 85.12$\pm$0.80 & 84.76$\pm$0.84
& 57.19$\pm$0.64 & 54.66$\pm$0.31
& 71.32$\pm$1.62 & 70.28$\pm$1.35
& 71.12$\pm$0.32 & 70.47$\pm$0.52
& 78.27$\pm$0.69 & 78.05$\pm$0.70 \\

CTNet
& 86.27$\pm$0.63 & 85.76$\pm$0.84
& 57.69$\pm$0.71 & 55.95$\pm$0.60
& \underline{74.19$\pm$1.15} & \underline{73.57$\pm$1.37}
& 72.13$\pm$0.44 & 71.38$\pm$0.61
& 79.02$\pm$0.39 & 78.67$\pm$0.49 \\

TMSA-Net
& 84.92$\pm$0.33 & 84.32$\pm$0.28
& 56.41$\pm$0.39 & 55.55$\pm$0.60
& 71.47$\pm$1.32 & 70.88$\pm$1.07
& 70.37$\pm$0.46 & 69.83$\pm$0.55
& 79.85$\pm$0.46 & 79.63$\pm$0.46 \\

MGFormer
& 86.31$\pm$0.70 & 86.09$\pm$0.68
& 55.66$\pm$0.34 & 53.36$\pm$0.57
& 72.85$\pm$0.87 & 72.00$\pm$0.80
& 70.85$\pm$0.25 & 70.08$\pm$0.11
& 78.25$\pm$0.71 & 77.90$\pm$0.75 \\

Deformer
& 85.91$\pm$0.62 & 85.36$\pm$0.85
& \underline{57.74$\pm$0.57} & 54.20$\pm$0.57
& 70.83$\pm$1.52 & 69.90$\pm$1.96
& \underline{72.53$\pm$1.48} & \underline{71.59$\pm$1.63}
& \underline{82.35$\pm$0.39} & \underline{82.04$\pm$0.42} \\

MSCFormer
& 86.52$\pm$0.39 & 86.11$\pm$0.36
& 57.33$\pm$0.93 & 56.16$\pm$1.27
& 73.59$\pm$1.11 & 73.07$\pm$1.22
& 71.62$\pm$0.47 & 71.11$\pm$0.53
& 80.24$\pm$0.61 & 79.97$\pm$0.64 \\

\hline
\textbf{DSAINet (Ours)}
& \textbf{89.33$\pm$0.72} & \textbf{89.22$\pm$0.78}
& \textbf{59.65$\pm$0.55} & \textbf{58.40$\pm$0.92}
& \textbf{75.85$\pm$1.27} & \textbf{75.52$\pm$1.18}
& \textbf{72.83$\pm$0.28} & \textbf{72.03$\pm$0.53}
& \textbf{84.81$\pm$0.52} & \textbf{84.67$\pm$0.51} \\

\hline
\end{tabular}
\end{table*}

Tables~\ref{tab:mi_result} and~\ref{tab:other_result} summarize the performance of the compared methods on ten EEG classification datasets. Overall, DSAINet achieves the best ACC and weighted F1-score on all datasets under subject-independent settings, showing that the proposed dual-scale interactive attention design generalizes well across five MI datasets and five heterogeneous EEG datasets from other paradigms.

The five MI datasets cover diverse settings in subject number, channel number, and class number, providing a broad evaluation of model robustness. They include limited-subject and more difficult settings (Zhou2016 and BCIC-IV-2a), an extremely low-channel setting (BCIC-IV-2b), a larger-subject setting (OpenBMI), and a high-channel, large-subject setting with broader spatial coverage beyond the motor cortex (PhysioNet-MI). Across these variations, DSAINet maintains the best ACC and weighted F1-score, highlighting its robustness under diverse experimental conditions. The gains are especially evident on BCIC-IV-2a and Zhou2016, where DSAINet improves ACC by 1.71\% and 1.52\% over the second-best methods, respectively, while also improving weighted F1-score by 2.03\% and 1.56\%. DSAINet also shows consistent gains on OpenBMI and PhysioNet-MI, demonstrating robustness from relatively compact to larger and higher-channel MI settings. On BCIC-IV-2b, it attains the best ACC and a tie for the best weighted F1-score despite the extremely limited number of channels, indicating that its effectiveness does not rely on rich spatial coverage. Moreover, DSAINet shows low standard deviations across all MI datasets, indicating stable cross-fold or cross-subject generalization rather than gains driven by a few favorable splits.

Consistent gains are also observed on the five datasets from four other EEG decoding tasks, further showing that the advantage of DSAINet is not limited to MI under subject-independent evaluation. Notably, DSAINet ranks first on all five datasets in terms of both ACC and weighted F1-score. The most pronounced improvement is achieved on Shin2018, where DSAINet exceeds the second-best method by 2.46\% in ACC and 2.63\% in weighted F1-score. Clear gains are also observed on ADFTD, with improvements of 1.91\% in ACC and 1.80\% in weighted F1-score. On Mumtaz2017, Rockhill2021 and EEGMat, DSAINet continues to deliver the best overall results with consistent improvements over competing methods. These results suggest that the same architecture configuration remains effective across heterogeneous EEG decoding scenarios with different task properties, channel configurations, and temporal characteristics.

\subsection{Ablation Study}
Table~\ref{tab:ablation_single} presents the ablation results on three representative tasks, including two MI datasets (BCIC-IV-2a and BCIC-IV-2b) and one neurodegenerative disorder dataset (ADFTD). In all three cases, the full DSAINet achieves the highest ACC, indicating that each component contributes positively to the final model. The consistent performance drops observed after removing individual components further suggest that the overall improvement stems from the coordinated effect of Shared Spatiotemporal Tokenization, Fine/Coarse-Scale Temporal Convolution, Intra-Branch Attentive Refinement, Inter-Branch Attentive Interaction, and Adaptive Token Aggregation, rather than from any single module in isolation.

\begin{table}[!t]
\caption{Ablation study of DSAINet on three representative tasks using ACC (\%), reported as mean$\pm$std.}
\label{tab:ablation_single}
\centering
\renewcommand{\arraystretch}{1.3}
\setlength{\tabcolsep}{2pt}
\begin{tabular}{l|c|c|c}
\hline
\centering Model Variant
& \multicolumn{1}{c|}{BCIC-IV-2a}
& \multicolumn{1}{c|}{BCIC-IV-2b}
& \multicolumn{1}{c}{ADFTD} \\
\hline

w/o Positional Embedding
& 59.76$\pm$1.26
& 75.71$\pm$0.97
& 57.06$\pm$0.87 \\

Single Fine-Scale Branch
& 59.82$\pm$0.58
& 75.94$\pm$0.57
& 58.57$\pm$0.46 \\

Single Coarse-Scale Branch
& 60.19$\pm$1.02
& 75.74$\pm$0.80
& 57.84$\pm$1.41 \\

w/o Attentive Interaction 
& 59.36$\pm$0.61
& 75.90$\pm$0.54
& 57.62$\pm$1.19 \\

w/o Intra-Branch Attn. 
& 60.07$\pm$0.63
& 75.49$\pm$0.51
& 58.97$\pm$1.40 \\

w/o Inter-Branch Attn. 
& 60.14$\pm$0.49
& 75.56$\pm$0.69
& 57.65$\pm$1.51 \\

w/o Adaptive Token Aggregation 
& 59.82$\pm$0.74
& 75.58$\pm$0.45
& 56.92$\pm$1.20 \\

\hline
\textbf{DSAINet (Full Model)}
& \textbf{61.79$\pm$0.68}
& \textbf{76.41$\pm$0.31}
& \textbf{59.65$\pm$0.55} \\

\hline
\end{tabular}
\end{table}

\subsubsection{Impact of Positional Embedding}

Removing positional embedding reduces ACC on all three tasks, from 61.79\% to 59.76\% on BCIC-IV-2a, from 76.41\% to 75.71\% on BCIC-IV-2b, and from 59.65\% to 57.06\% on ADFTD. This indicates that, after Shared Spatiotemporal Tokenization and temporal convolution, preserving token order remains important for the subsequent attentive modules. Explicit positional cues help maintain discriminative temporal structure by telling the model not only what patterns appear, but also where they occur in the token sequence.

\subsubsection{Impact of Fine/Coarse-Scale Temporal Convolution}

Replacing Fine/Coarse-Scale Temporal Convolution with a single-branch variant consistently reduces ACC. On BCIC-IV-2a, the single fine-scale and single coarse-scale variants achieve 59.82\% and 60.19\% ACC, both below the full model, and the same trend is observed on BCIC-IV-2b and ADFTD. These results show that neither local nor broader temporal modeling alone is sufficient. Instead, combining fine- and coarse-scale branches enables the model to learn richer multi-scale temporal features.

\subsubsection{Impact of Intra-Branch Attentive Refinement and Inter-Branch Attentive Interaction}

Removing attentive interaction between branches also degrades ACC. When the two branches are fused only at a late stage, ACC drops to 59.36\% on BCIC-IV-2a, 75.90\% on BCIC-IV-2b, and 57.62\% on ADFTD. Ablating Intra-Branch Attentive Refinement or Inter-Branch Attentive Interaction likewise reduces performance. The relative drop is not identical across tasks: some tasks appear to rely more on scale-specific refinement within each branch, whereas others benefit more from explicit interaction across scales. This is consistent with the fact that different EEG tasks may prefer different temporal extents and different degrees of cross-scale integration. Nevertheless, the full design remains consistently superior, showing that the two branches should not only coexist, but also interact adaptively.

\begin{figure*}[t]
\centering
\includegraphics[width=\textwidth,trim={0 0 0 0},clip]{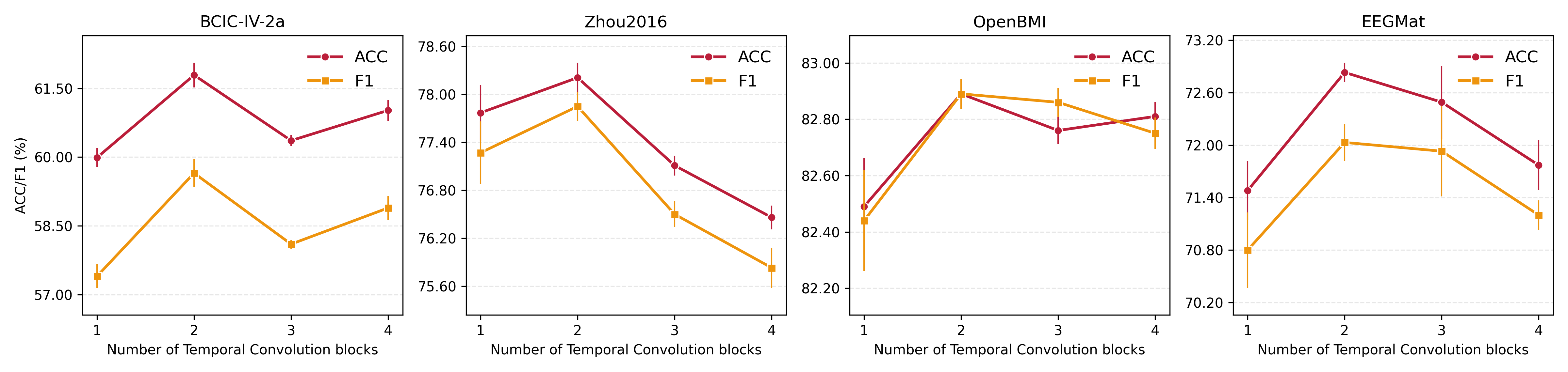}
\includegraphics[width=\textwidth,trim={0 0 0 0},clip]{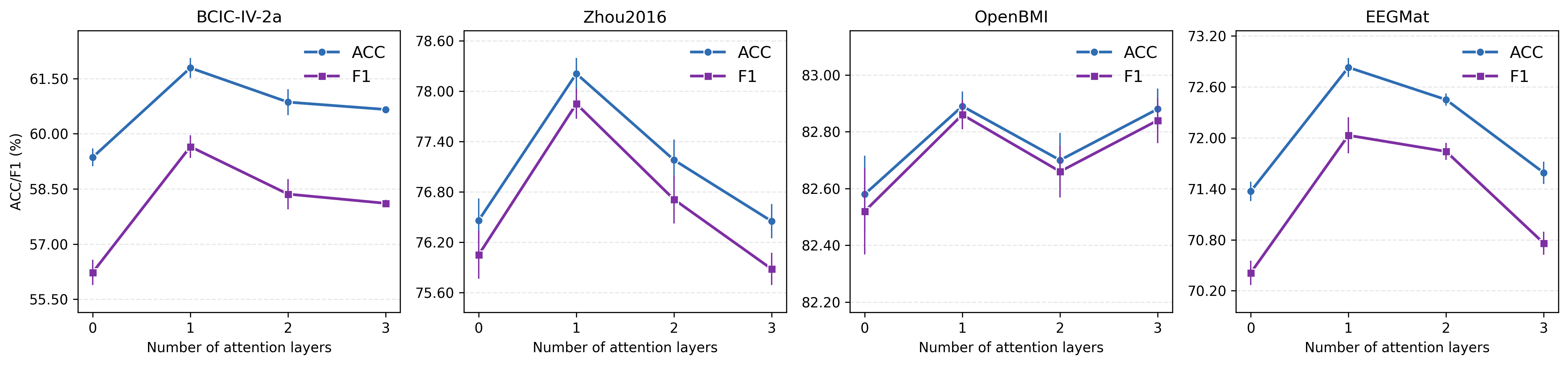}
\captionsetup{skip=2pt}
\caption{Impact of architectural hyperparameters on classification performance on BCIC-IV-2a, Zhou2016, OpenBMI, and EEGMat. The first row illustrates the effect of the number of Temporal Convolution blocks in each branch. The second row shows the effect of attention depth.}
\label{fig:depth_width_ablation}
\end{figure*}

\subsubsection{Impact of Adaptive Token Aggregation}

Replacing Adaptive Token Aggregation with simple average pooling also reduces ACC, with the largest drop observed on ADFTD (from 59.65\% to 56.92\%). This indicates that EEG tokens are not equally informative over time. An adaptive aggregation mechanism can assign larger weights to task-relevant tokens and suppress less informative ones, leading to more discriminative trial-level representations.

\subsection{Hyperparameter Sensitivity Analysis}

On four representative datasets, Fig.~\ref{fig:depth_width_ablation} shows the impact of two architectural hyperparameters: the number of Temporal Convolution blocks in each branch and the number of attentive layers. DSAINet remains fairly stable under moderate changes, with the best overall setting obtained using two Temporal Convolution blocks and one attentive layer, which is adopted in the main experiments.

\subsubsection{Effect of the number of Temporal Convolution Blocks per Branch}

As shown in the first row of Fig.~\ref{fig:depth_width_ablation}, increasing the number of Temporal Convolution blocks from one to two consistently improves performance on all four datasets. For example, ACC on BCIC-IV-2a increases from 59.99\% to 61.79\%, and ACC on EEGMat rises from 71.48\% to 72.83\%. Similar gains are observed on Zhou2016 and OpenBMI. However, further increasing the depth to three or four blocks brings little additional benefit and may even cause a slight decline. This suggests that moderate temporal depth is sufficient to capture task-relevant multi-scale temporal features, whereas deeper Temporal Convolution stacks may introduce redundancy and make optimization harder. Therefore, two Temporal Convolution blocks per branch provide a good balance between capacity and stability across datasets.

\subsubsection{Effect of the number of Attentive Layers}

A similar pattern is observed for attentive depth. In most cases, one attentive layer already gives the best or near-best performance, while stacking more layers brings limited or negative returns. This suggests that, once Fine/Coarse-Scale Temporal Convolution has produced informative branch features, a shallow attentive stage is sufficient for Intra-Branch Attentive Refinement and Inter-Branch Attentive Interaction. Because these attentive modules have clearly separated roles, they do not need to be stacked deeply to handle more complex functions. Thus, the empirical results support a compact one-layer attentive design.

\subsection{Effect of Segment Length}

\begin{figure}[t]
    \centering
    \includegraphics[width=1.0\linewidth, trim={0 6 0 0},clip]{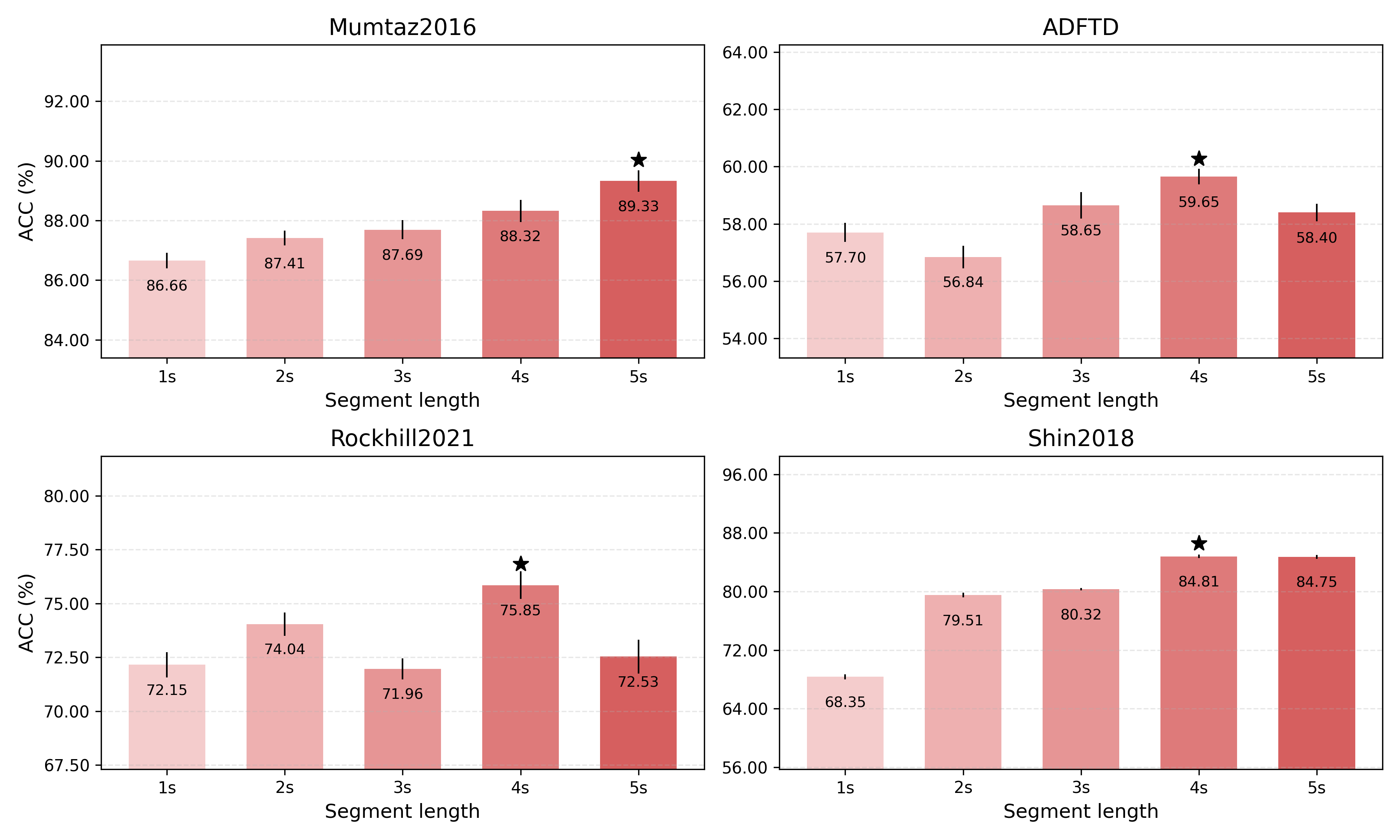}
    \captionsetup{skip=2pt}
    \caption{Effect of segment length on classification performance on Mumtaz2017, ADFTD, Rockhill2021, and Shin2018. Stars indicate the segment lengths adopted in the main experiments.}
    \label{fig:segment_length}
\end{figure}

Because several downstream datasets are constructed by segmenting longer EEG recordings into labeled samples, segment length affects both the temporal context retained in each input and the resulting sample construction. We therefore examine its effect to assess the robustness of DSAINet under different segmentation settings. As shown in Fig.~\ref{fig:segment_length}, DSAINet maintains strong performance across a range of segment lengths, with the best results generally obtained using moderate windows. Performance improves steadily with longer segments on Mumtaz2017, peaks at 4\,s on ADFTD and Rockhill2021, and rises markedly from 1\,s to 4\,s on Shin2018 with only marginal change thereafter. Overall, these results indicate that DSAINet adapts well to different temporal contexts without being overly sensitive to the exact segment length. The segment lengths adopted in the main experiments were therefore chosen as stable and effective operating points.

\subsection{Computational Complexity}

\begin{figure}[t]
    \centering
    \includegraphics[width=1\linewidth, trim={0 4 0 0},clip]{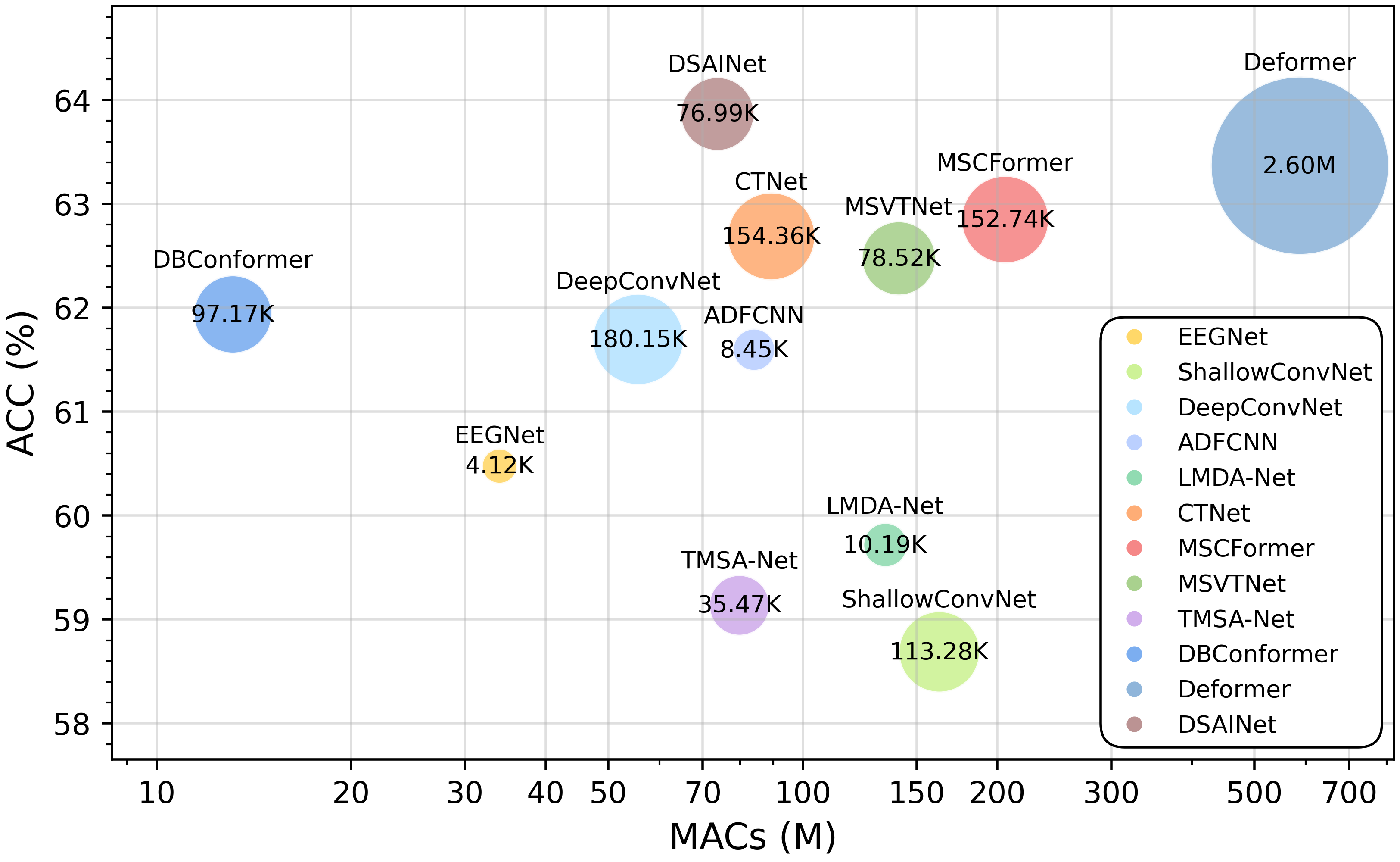}
    \captionsetup{skip=2pt}
    \caption{Accuracy-efficiency trade-off on the PhysioNet-MI dataset. The x-axis shows computational cost in MACs (M) on a logarithmic scale, and the y-axis shows ACC (\%). Bubble area is proportional to the number of trainable parameters, with exact parameter counts annotated.}
    \label{fig:efficiency}
\end{figure}

Fig.~\ref{fig:efficiency} compares the accuracy-efficiency trade-off on PhysioNet-MI. DSAINet achieves the highest ACC among all compared models at 63.87\%, while remaining substantially more efficient than the strongest competing baselines. In particular, compared with Deformer, DSAINet improves ACC from 63.37\% to 63.87\% while reducing MACs from 586.92 MMACs to 73.67 MMACs and parameters from 2.60M to 76.99K. Compared with other strong CNN-Transformer hybrid models, such as CTNet, MSCFormer, and MSVTNet, DSAINet also achieves higher accuracy with lower parameter cost and competitive computational cost. Overall, these results place DSAINet in a favorable accuracy-efficiency region, highlighting its practicality for EEG decoding scenarios where both predictive performance and efficiency are important.

\subsection{Interpretability and Visualization}

\subsubsection{Saliency Maps Visualization}

\begin{figure}[t]
\centering

\begin{minipage}[t]{0.30\linewidth}
    \centering
    \includegraphics[width=\linewidth]{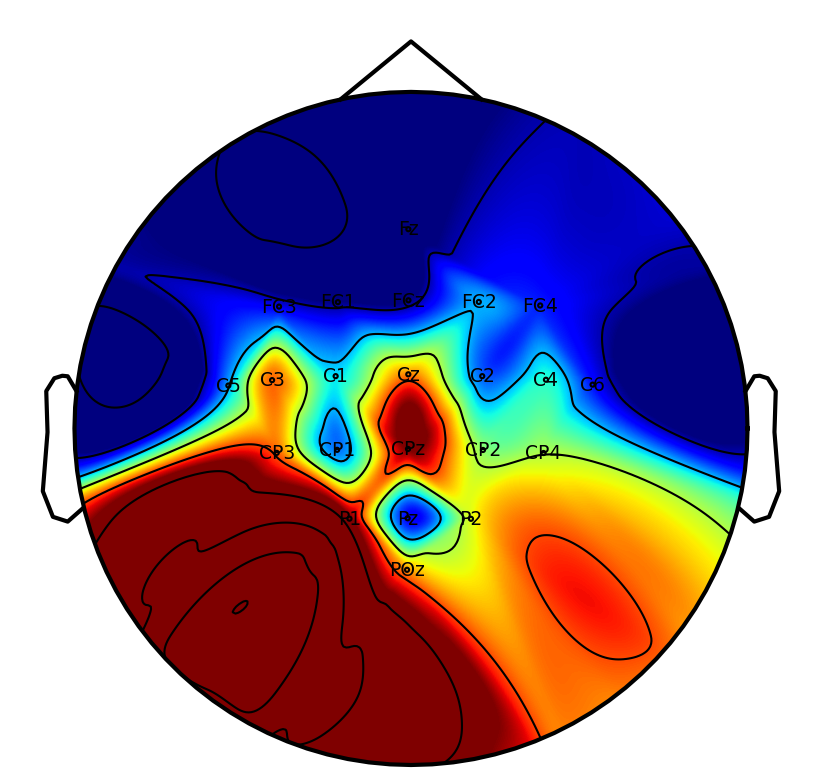}\\
    \small BCIC-IV-2a
\end{minipage}\hfill
\begin{minipage}[t]{0.30\linewidth}
    \centering
    \includegraphics[width=\linewidth]{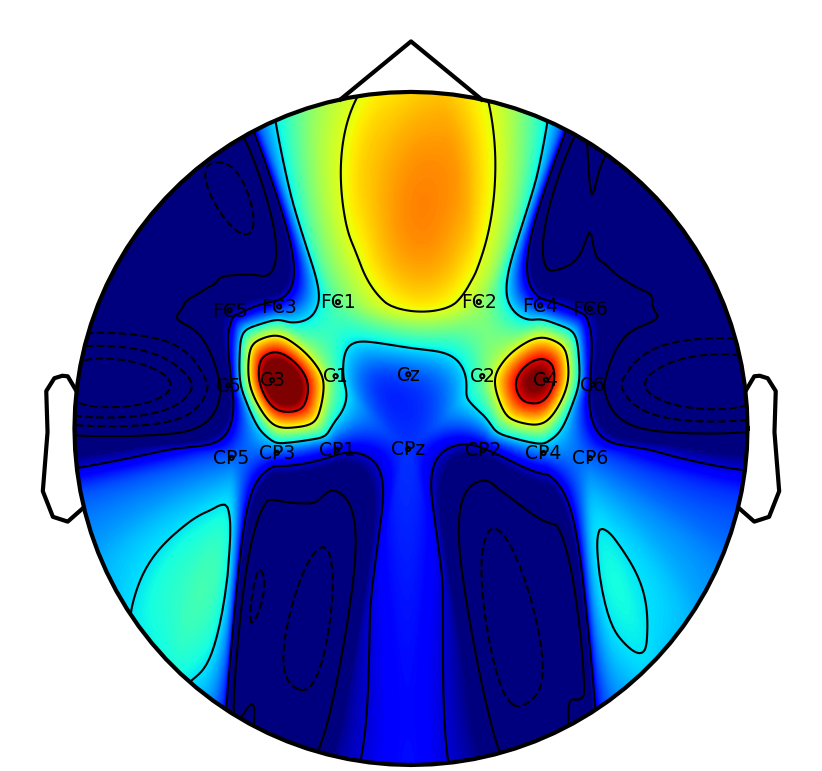}\\
    \small OpenBMI
\end{minipage}\hfill
\begin{minipage}[t]{0.30\linewidth}
    \centering
    \includegraphics[width=\linewidth]{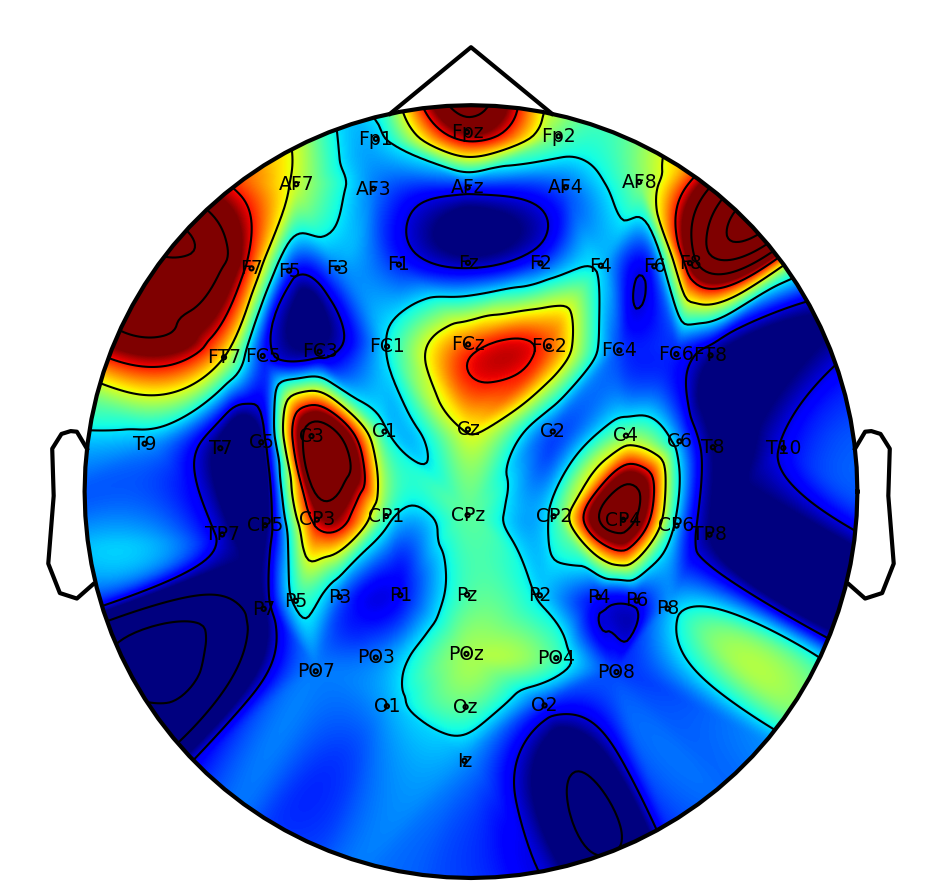}\\
    \small PhysioNet-MI
\end{minipage}

\vspace{0.6em}

\begin{minipage}[t]{0.30\linewidth}
    \centering
    \includegraphics[width=\linewidth]{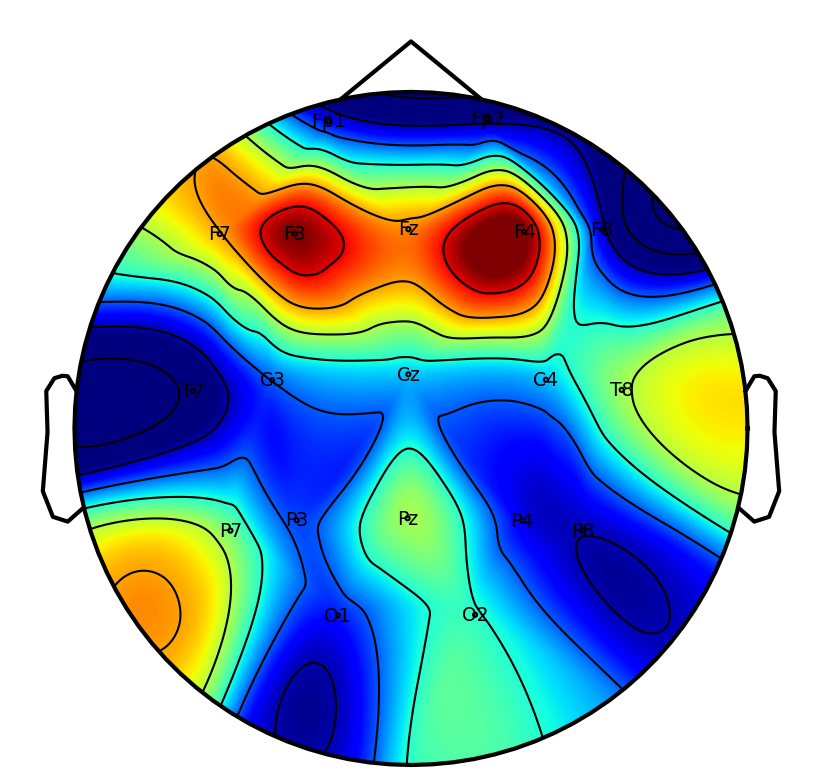}\\
    \small Mumtaz2017
\end{minipage}\hfill
\begin{minipage}[t]{0.30\linewidth}
    \centering
    \includegraphics[width=\linewidth]{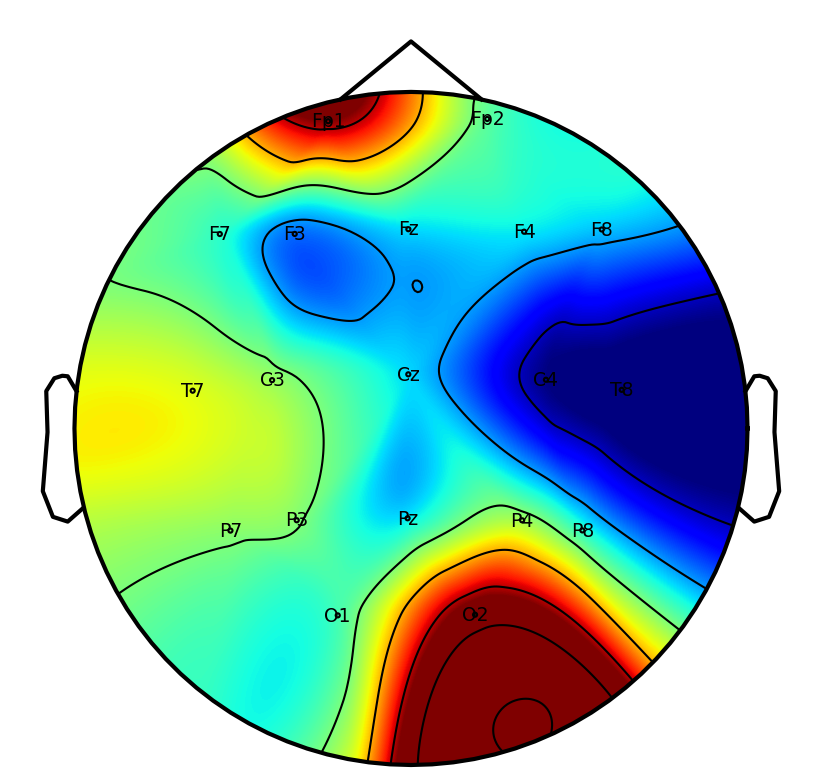}\\
    \small ADFTD-AD
\end{minipage}\hfill
\begin{minipage}[t]{0.30\linewidth}
    \centering
    \includegraphics[width=\linewidth]{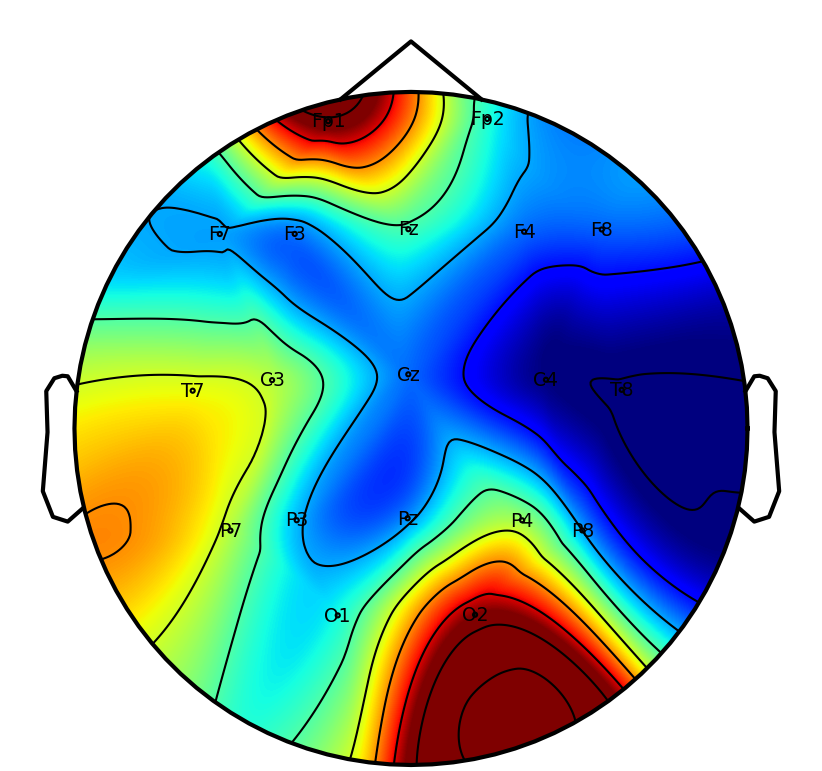}\\
    \small ADFTD-FTD
\end{minipage}

\vspace{0.6em}

\begin{minipage}[t]{0.30\linewidth}
    \centering
    \includegraphics[width=\linewidth]{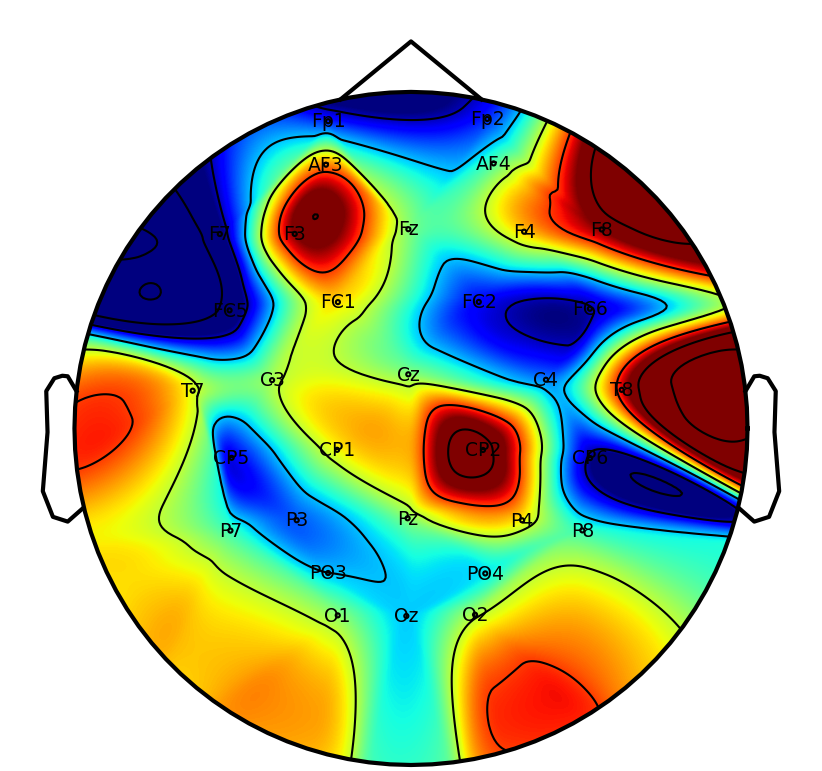}\\
    \small Rockhill2021
\end{minipage}\hfill
\begin{minipage}[t]{0.30\linewidth}
    \centering
    \includegraphics[width=\linewidth]{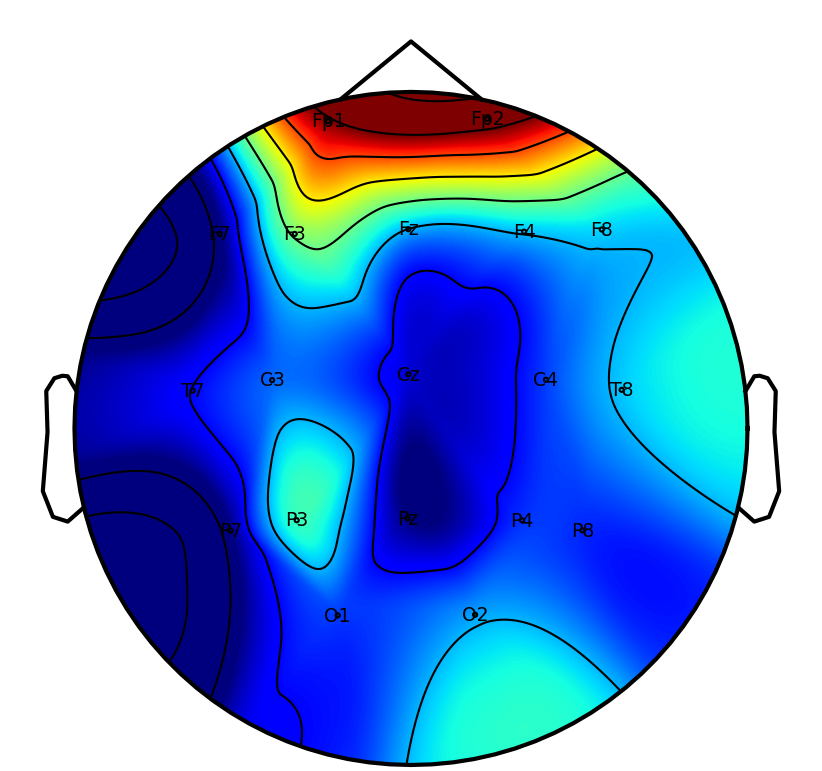}\\
    \small EEGMat
\end{minipage}\hfill
\begin{minipage}[t]{0.30\linewidth}
    \centering
    \includegraphics[width=\linewidth]{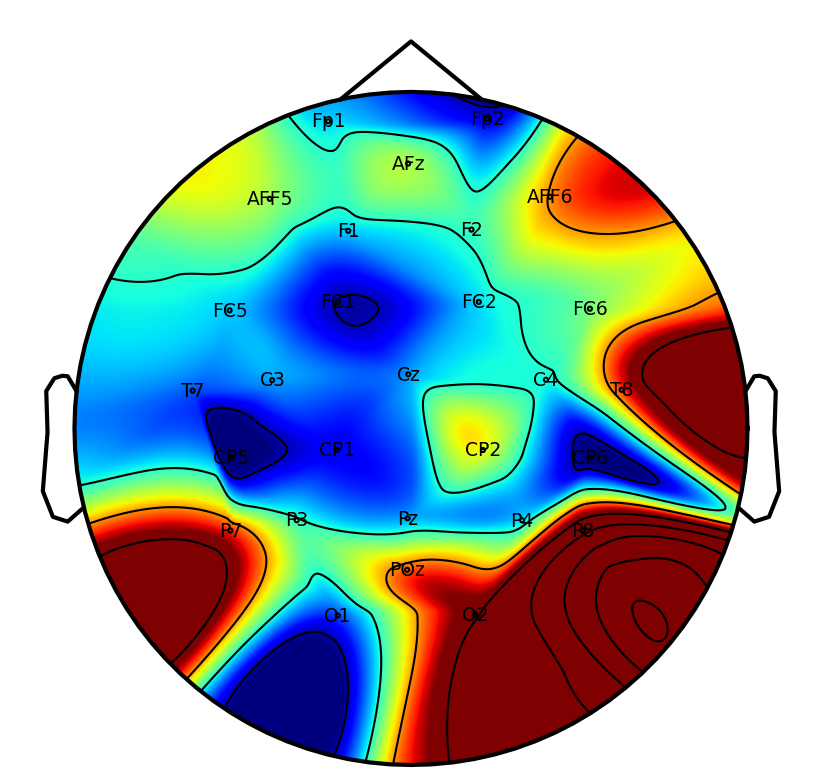}\\
    \small Shin2018
\end{minipage}

\captionsetup{skip=6pt}
\caption{Mean saliency maps averaged across all folds (or across all subjects for BCIC-IV-2a). Nine saliency maps from eight representative datasets are shown. For disorder-related datasets, only subjects from the corresponding patient groups are included, i.e., MDD patients for Mumtaz2017, PD patients for Rockhill2021, and AD/FTD patients separately for ADFTD.}
\label{fig:saliency_maps}
\end{figure}

\begin{figure*}[t]
\centering



\includegraphics[width=0.96\textwidth]{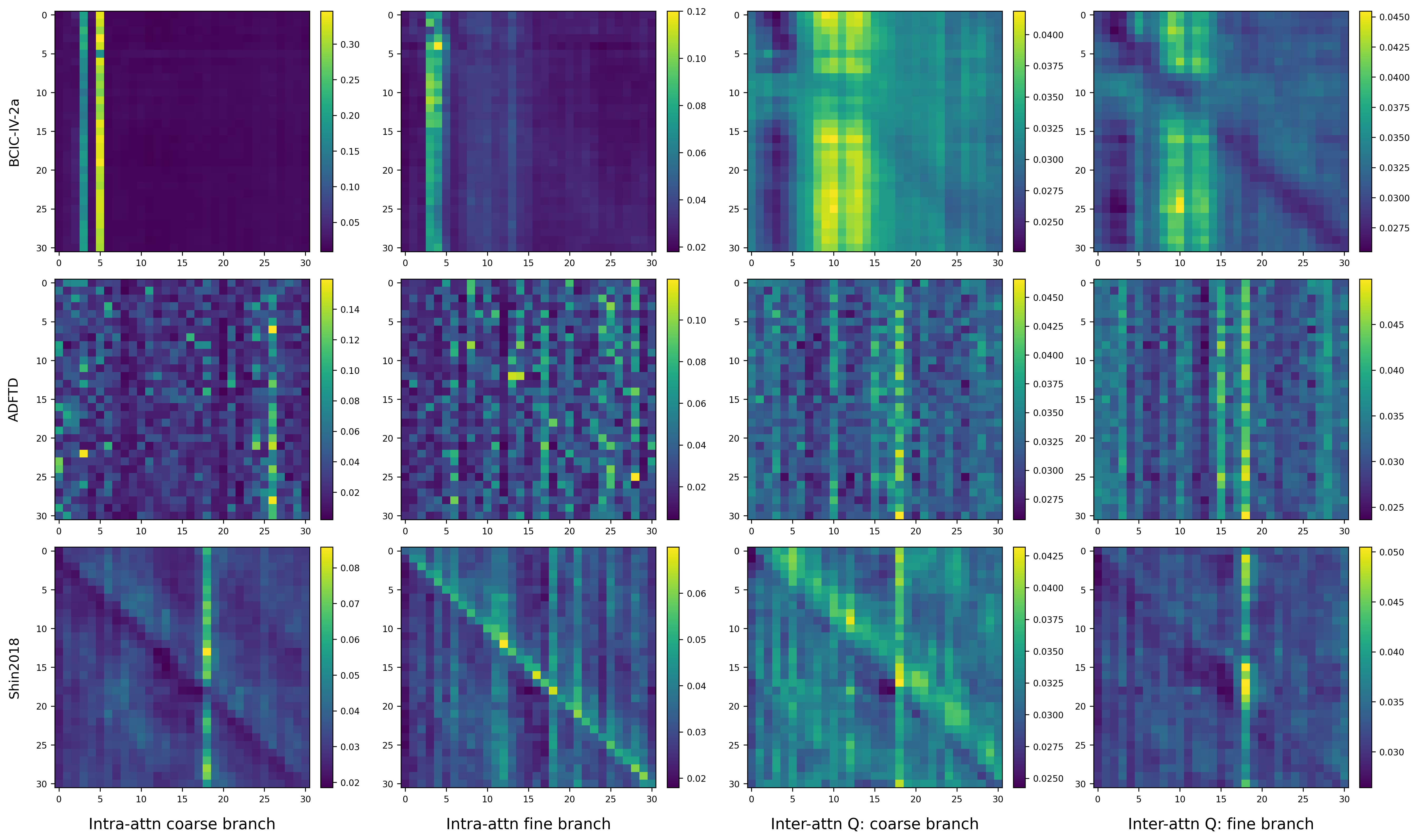}

\caption{Learned attention patterns of Intra-Branch Attentive Refinement and Inter-Branch Attentive Interaction on BCIC-IV-2a, ADFTD, and Shin2018. Each row shows two intra-branch attention maps and two inter-branch attention maps, illustrating dataset-dependent yet structured branch refinement and cross-branch feature interaction.}
\label{fig:attn_patterns}
\end{figure*}
The saliency maps in Fig.~\ref{fig:saliency_maps} show clear dataset-specific yet spatially organized patterns, suggesting that DSAINet adapts to task-relevant neural signatures rather than relying on a fixed spatial bias.

For the MI datasets, the saliency maps are concentrated over sensorimotor regions. In BCIC-IV-2a, the strongest responses lie around \textit{C3}, \textit{Cz}, \textit{C4}, and adjacent sites such as \textit{CP3}, \textit{CPz}, and \textit{CP4}, consistent with the lateralized ERD/ERS effects over sensorimotor cortex during motor imagery \cite{pfurtscheller2001motor}. OpenBMI shows especially sharp bilateral peaks at \textit{C3} and \textit{C4}, matching the enlarged motor-cortex coverage in this dataset. In PhysioNet-MI, salient regions remain centered on \textit{C3}/\textit{CP3} and \textit{C4}/\textit{CP4}, with additional frontal emphasis around \textit{FCz}, \textit{FC2}, \textit{F7}, and \textit{F8}, which is compatible with frontal motor planning during imagery \cite{decety1996neurophysiological}.

For the mental disorder dataset Mumtaz2017, the map is dominated by frontal electrodes, with clear peaks around \textit{F3}, \textit{F4}, and \textit{Fz}. These prefrontal regions are closely related to depression discrimination, and similar frontal abnormalities have been reported in EEG-based depression analysis \cite{peng2019multivariate}. For the neurodegenerative disorder dataset ADFTD, salient regions appear in both anterior and posterior cortex. The AD and FTD maps both emphasize \textit{Fp1} and posterior right-sided regions around \textit{O2}, \textit{P4}, and \textit{P8}, while the FTD map shows relatively stronger left temporal involvement near \textit{T7}. Such anterior-posterior differences are consistent with disease-related cortical dysfunction in dementia \cite{iizuka2019deep}. For the neurodegenerative disorder dataset Rockhill2021, DSAINet attends to distributed frontal and centroparietal areas, with prominent responses around \textit{AF3}/\textit{F3}, \textit{F4}/\textit{F8}, and \textit{CP2}/\textit{P4}, which agrees with Parkinson-related alterations in frontal executive and sensorimotor processing \cite{han2013investigation}.

For the mental workload dataset EEGMat, the saliency map is mainly frontal, with strongest emphasis at \textit{Fp1} and \textit{Fp2}, extending toward \textit{F3}/\textit{Fz}, together with weaker parietal involvement near \textit{P3} and \textit{P4}. This is consistent with prior findings that frontal and parietal regions are informative for workload assessment \cite{kakkos2021eeg}. For the cognitive attention dataset Shin2018, salient regions span frontal and posterior-parietal areas, particularly \textit{AFF5}, \textit{AFF6}, \textit{AFz}, \textit{F1}/\textit{F2}, and posterior sites such as \textit{P7}, \textit{P4}, \textit{POz}, and \textit{O2}. This fronto-parietal organization agrees with cortical networks involved in attention and target selection \cite{liu2016top}. Overall, the saliency maps remain task-consistent within each dataset while differing across datasets, supporting that DSAINet learns physiologically meaningful and task-dependent spatial representations.

\subsubsection{Learned Attention Patterns}

To better understand the distinct roles of the two attentive modules, we visualize the learned attention maps of Intra-Branch Attentive Refinement and Inter-Branch Attentive Interaction on three representative datasets, namely BCIC-IV-2a, ADFTD, and Shin2018. As shown in Fig.~\ref{fig:attn_patterns}, these maps represent pairwise attention weights between feature tokens. The maps are highly structured but differ noticeably across datasets, suggesting that the proposed dual-branch design does not rely on a fixed interaction pattern. Overall, the model adaptively organizes branch-wise refinement and cross-branch feature interaction according to the temporal characteristics of each task.

For BCIC-IV-2a, the intra-branch attention maps are relatively concentrated, while the inter-branch attention maps spread their responses over broader token regions. This indicates that Intra-Branch Attentive Refinement mainly sharpens scale-specific patterns, whereas Inter-Branch Attentive Interaction further integrates features from different time scales. For ADFTD, the intra-branch attention maps appear comparatively sparse, while the inter-branch attention maps reveal clearer diffused dependencies, suggesting that cross-scale interaction is particularly important for disorder-related resting-state EEG. For Shin2018, the attention maps exhibit more distributed structures with clear diagonal and band-like patterns, implying the coexistence of short-range token continuity and longer-range interactions. Overall, these results support the intended roles of the two attention mechanisms: Intra-Branch Attentive Refinement performs branch-specific refinement, while Inter-Branch Attentive Interaction promotes complementary fusion between fine- and coarse-scale representations.

\section{Limitations and Future Work}

Despite the strong overall performance of DSAINet, several limitations remain. First, to preserve the shared architectural configuration required for general EEG decoding, the current framework adopts a largely task-agnostic Shared Spatiotemporal Tokenization and attention design, and therefore does not explicitly exploit EEG-specific priors such as electrode topology, anatomical structure, or frequency-dependent characteristics. Second, although DSAINet achieves a favorable accuracy-efficiency trade-off, its practical efficiency has so far been assessed mainly through offline complexity comparisons rather than deployment-oriented evaluation. Third, while the present results demonstrate strong cross-task architectural generality under subject-independent evaluation, they should still be interpreted as evidence for a unified and reusable decoding framework across the evaluated settings, rather than a definitive solution for all EEG scenarios.

Future work will therefore focus on three directions. One is to develop more structure-aware variants of DSAINet by incorporating EEG-specific priors while retaining the shared backbone. Another is to further improve deployment efficiency through compression, pruning, or distillation for low-latency EEG decoding. The third is to evaluate DSAINet on a broader range of paradigms and larger cross-subject benchmarks, while further investigating its task-adaptive multi-scale temporal modeling behavior across tasks.

\section{Conclusion}

In this study, we propose DSAINet for general EEG decoding under strict subject-independent evaluation. Extensive experiments on five EEG paradigms across ten public datasets show that DSAINet consistently delivers strong performance with a shared architectural configuration across tasks, while maintaining a favorable accuracy-efficiency trade-off. This effectiveness is supported by a clearly structured design, in which Shared Spatiotemporal Tokenization provides a common backbone for diverse EEG inputs, and Fine/Coarse-Scale Temporal Convolution together with the two attentive modules enables complementary temporal information to be refined within each scale and integrated across scales. Ablation and interpretability analyses further suggest that its advantage extends across heterogeneous EEG decoding scenarios with different task characteristics and recording conditions, and is supported by the coordinated effect of multi-scale temporal modeling, within-scale refinement, and cross-scale interaction. Taken together, these findings support DSAINet as an effective and reusable solution for subject-independent EEG decoding, and highlight the promise of compact unified architectures for handling diverse EEG tasks within a common framework.

\section*{References} 
\vspace{-1.5em}
\bibliographystyle{IEEEtran}
\bibliography{reference}
\end{document}